\documentclass[10pt,twocolumn,letterpaper]{article}
\usepackage{tra-preprint}
\usepackage[normalem]{ulem}

\usepackage{booktabs}
\usepackage[table]{xcolor}

\definecolor{reasonblue}{HTML}{E5EFF9}
\definecolor{stoporange}{HTML}{FCE8D5}

\title{Privileged Teachers Were Never About Teaching Knowledge:\\[1.5pt]
They Were Thinking Switches All Along}
\author{%
  Qiangqiang He\ensuremath{^{1,4,*,\dagger}}
  \qquad
  Jin Li\ensuremath{^{2,*}}
  \qquad
  Mingxuan Du\ensuremath{^{3, 4}}
  \qquad
  Benfeng Xu\ensuremath{^{3,4}}%
}
\date{}
\renewcommand{\trarunningtitle}{Privileged Teachers Were Never About Teaching Knowledge: They Were Thinking Switches All Along}
\renewcommand{\tracoreclaim}{I think, therefore I am.}
\hypersetup{
  pdftitle={Privileged Teachers Were Never About Teaching Knowledge: They Were Thinking Switches All Along},
  pdfauthor={Qiangqiang He, Jin Li, Mingxuan Du, and Benfeng Xu},
  pdfsubject={Teacher Reversal Alignment and Thinking-Controlled Policy Optimization},
  pdfkeywords={on-policy self-distillation, privileged teachers, thinking control, TRA, TCPO}
}

\usepackage{xcolor}
\usepackage{etoolbox}
\usepackage{ragged2e}
\usepackage[skins]{tcolorbox}

\usepackage{booktabs}
\usepackage{multirow}
\usepackage[table]{xcolor}

\definecolor{bestbg}{HTML}{E0F0E7}
\newcommand{\bestscore}[1]{\cellcolor{bestbg}\textbf{#1}}

\definecolor{keynoteBlue}{HTML}{244F73}
\definecolor{keynoteInk}{HTML}{172331}

\newtcolorbox{keynote}[1]{
  enhanced,
  width=\linewidth,
  colback=white,
  colframe=keynoteBlue!40!white,
  coltext=keynoteInk,
  coltitle=keynoteBlue,
  boxrule=0.5pt,
  arc=3pt,
  boxsep=0pt,
  left=10pt,
  right=10pt,
  top=13pt,
  bottom=10pt,
  before skip=10pt,
  after skip=12pt,
  fontupper=\normalfont\normalsize,
  fonttitle=\normalfont\bfseries\small,
  before upper={
    \setlength{\parindent}{0pt}
    \setlength{\parskip}{5pt}
    \RaggedRight
  },
  title={
    \textsc{Key Note}
    \ifstrempty{#1}{}{\enspace---\enspace#1}
  },
  attach boxed title to top left={
    xshift=8pt,
    yshift=-2mm
  },
  boxed title style={
    colback=white,
    colframe=white,
    boxrule=0pt,
    boxsep=0pt,
    left=4pt,
    right=4pt,
    top=1pt,
    bottom=1pt
  }
}

\begin{document}
\twocolumn[{\begin{@twocolumnfalse}
\maketitle
\begin{center}
  \fontsize{9.2}{12}\selectfont
  \textsuperscript{1}State Key Laboratory for Novel Software Technology, Nanjing University, Nanjing, China\\[2pt]
  \textsuperscript{2}College of Software Engineering, Southeast University, Nanjing, China\\[2pt]
  \textsuperscript{3}University of Science and Technology of China\\[2pt]
  \textsuperscript{4}MetaStone Technology\\[6pt]
  \href{mailto:qqh@smail.nju.edu.cn}{\texttt{qqh@smail.nju.edu.cn}}
  \qquad
  \href{mailto:jin_li@seu.edu.cn}{\texttt{jin\_li@seu.edu.cn}}\\[2pt]
  \href{mailto:dumingxuan@mail.ustc.edu.cn}{\texttt{dumingxuan@mail.ustc.edu.cn}}
  \qquad
  \href{mailto:benfeng@mail.ustc.edu.cn}{\texttt{benfeng@mail.ustc.edu.cn}}
\end{center}
\vspace{3pt}
\begin{preprintabstract}
On-Policy Self-Distillation (OPSD) is commonly interpreted as knowledge transfer from privileged teachers to students. We challenge this interpretation. Directly minimizing forward KL degrades reasoning performance and shortens responses, whereas the clipped forward-KL objective adopted by OPSD improves performance while increasing student--teacher KL divergence, revealing that its effective optimization drives students away from the teacher. We identify an underlying mechanism: alignment toward privileged teachers induces less thinking, whereas moving away induces more thinking. To validate this mechanism, we introduce \textbf{Teacher Reversal Alignment (TRA)}, which explicitly drives students away from privileged teachers. Across three Qwen3 models (1.7B, 4B, and 8B), TRA consistently achieves substantial improvements in reasoning performance alongside pronounced increases in reasoning length. Remarkably, these gains persist even when privileged solutions are completely irrelevant, demonstrating that task-specific solution knowledge is unnecessary for such improvements. Our findings reveal that privileged teachers are not knowledge providers but thinking switches: they control how much students think, rather than teaching them what to think. Our code is available at \url{https://github.com/qiangqiang-he/thinking-switches}.
\end{preprintabstract}

\vspace{8pt}

% Full-width figure fixed below the abstract.
\noindent\begin{minipage}{\textwidth}
    \centering
    \captionsetup{type=figure}

    \begin{subfigure}[t]{0.49\textwidth}
        \centering
        \includegraphics[height=4.7cm,keepaspectratio]
            {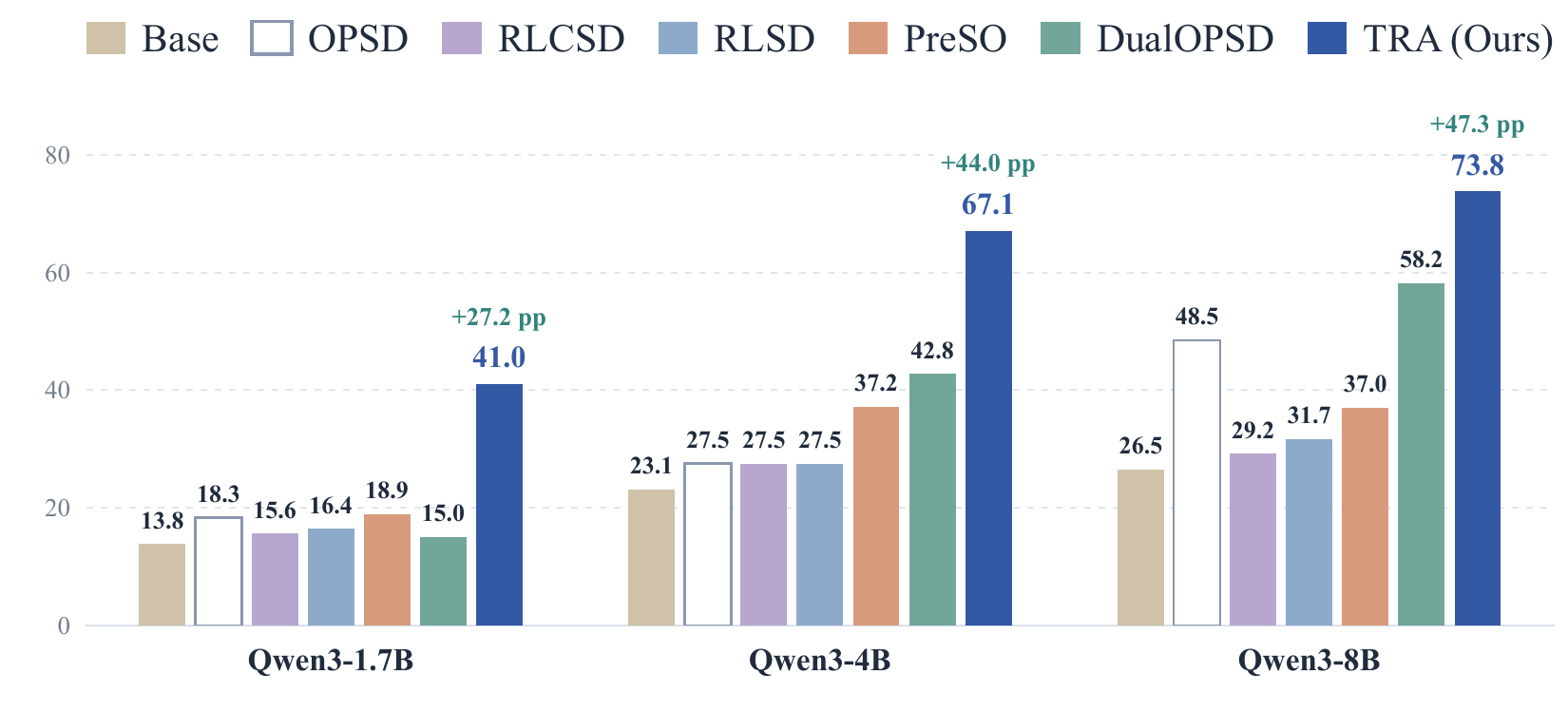}
        \caption{Reasoning performance (Avg@16).}
    \end{subfigure}\hfill
    \begin{subfigure}[t]{0.49\textwidth}
        \centering
        \includegraphics[height=4.5cm,keepaspectratio]
            {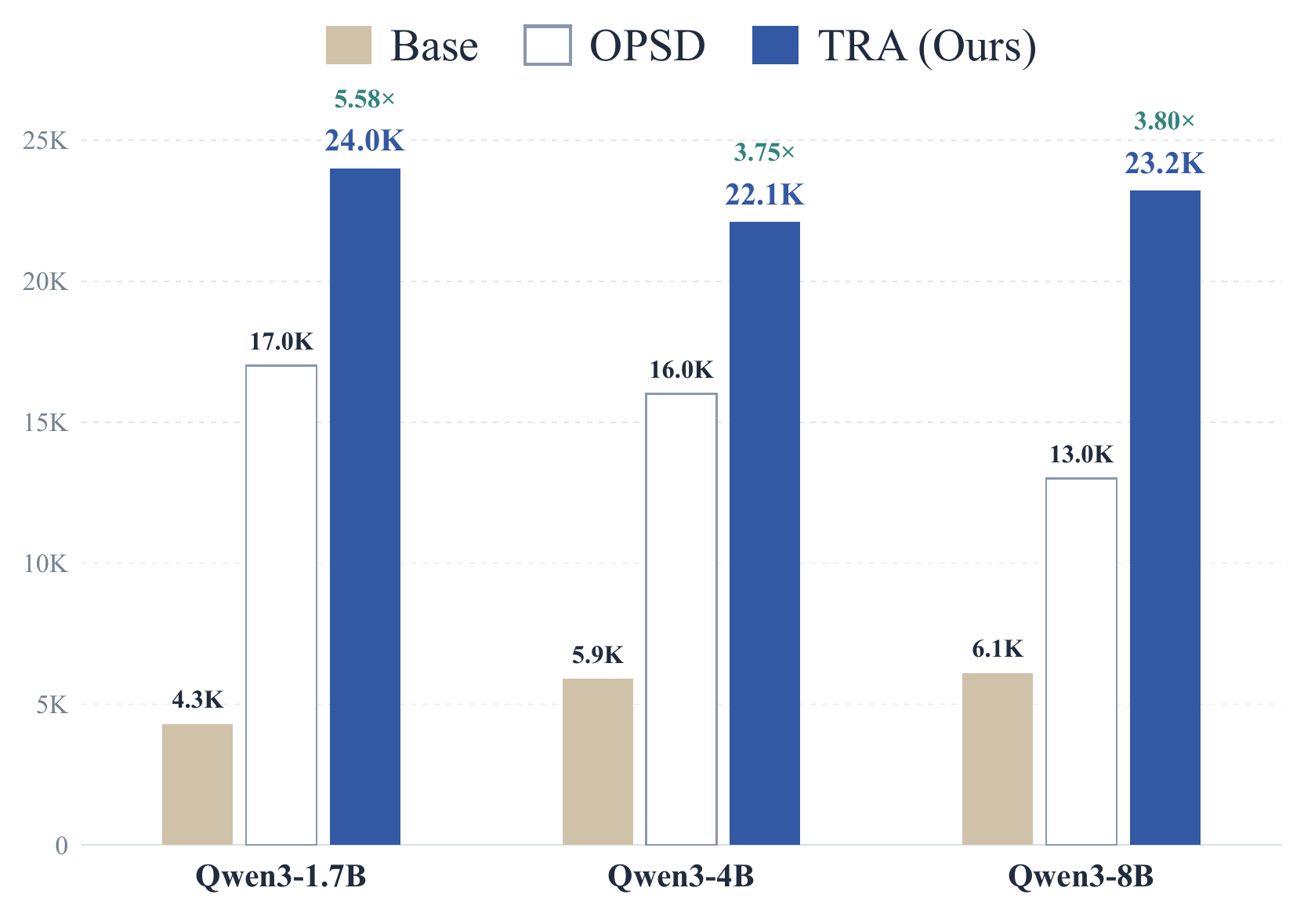}
        \caption{Average inference length.}
    \end{subfigure}

    \caption{AIME24 performance and reasoning length before and after TRA (32K-token limit). We compare TRA with OPSD~\citep{zhao2026self}, RLCSD~\citep{selfrlcsd}, RLSD~\citep{yang2026self}, PreSO~\citep{yang2026preso}, and DualOPSD~\citep{chen2026dualopsd}.}
    \label{fig:tra-aime24}
\end{minipage}
\par\vspace{8pt}

\end{@twocolumnfalse}}]

% Author notes must sit outside the full-width title block.
\begingroup
\renewcommand{\thefootnote}{\fnsymbol{footnote}}
\renewcommand{\footnotesize}{\fontsize{9}{11.2}\selectfont}
\setlength{\footnotesep}{8pt}

\footnotetext[1]{%
  \fontsize{9.5}{11.8}\selectfont\bfseries\color{traLink}\RaggedRight
  Equal contribution. Qiangqiang He and Jin Li are \mbox{co-first authors.}%
}

\footnotetext[2]{%
  \RaggedRight
  Work done during an internship at \mbox{MetaStone Technology.}%
}

\endgroup

\section{Introduction}
Knowledge distillation trains students to match teacher predictions, aiming to transfer the teacher's knowledge through distributional alignment~\citep{hinton2015distilling}. On-policy distillation extends this paradigm by aligning teacher and student distributions along trajectories sampled from the student itself~\citep{agarwal2024policy}. On-Policy Self-Distillation (OPSD) further extends this framework by using the same model as both teacher and student while providing privileged information, such as verified solutions, exclusively to the teacher~\citep{zhao2026self}. Under this formulation, privileged teachers are expected to transfer the knowledge contained in their additional context to students through token-level distributional supervision.

\begin{figure*}[t]
    \centering
    \includegraphics[width=\textwidth]{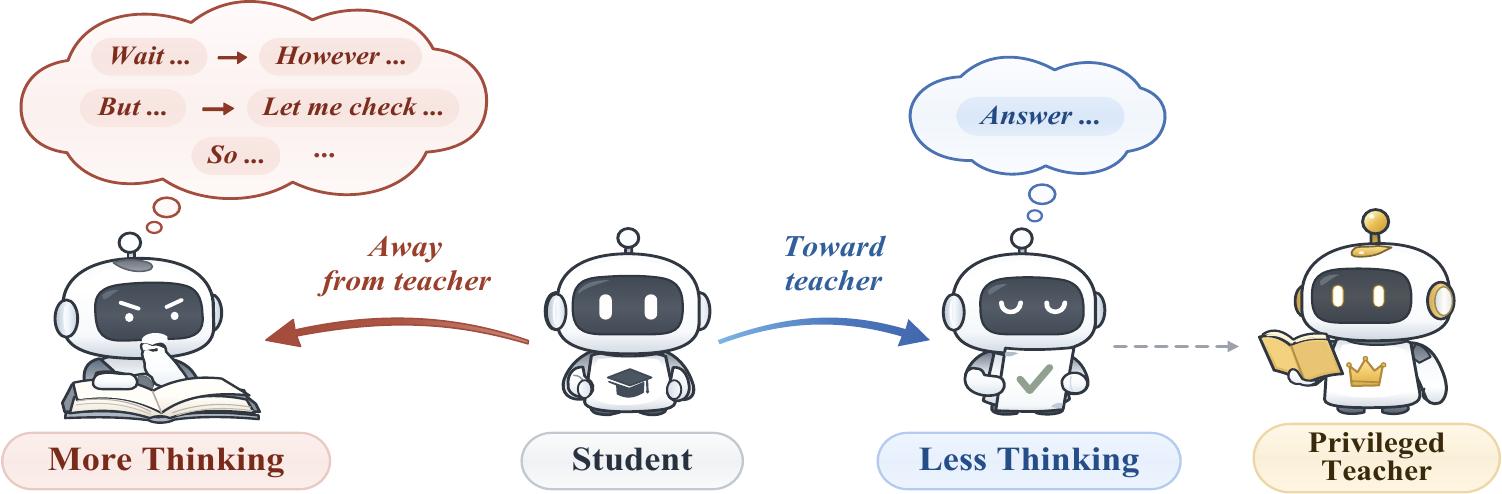}
    \caption{Privileged teachers are thinking switches: moving toward them reduces thinking, while moving away increases it.}
    \label{fig:student-teacher-thinking}
\end{figure*}

However, directly minimizing forward KL between student and privileged teacher distributions often degrades reasoning performance and reduces reasoning length~\citep{huang2026clipping, liu2026your, li2026diagnosing}. In contrast, OPSD adopts a clipped forward-KL objective~\citep{zhao2026self}, which can yield substantial performance gains. Yet, we uncover a striking discrepancy between its nominal objective and actual optimization dynamics: although the objective is intended to align the student with the privileged teacher, the student--teacher KL divergence increases steadily throughout training. In other words, the performance gains of OPSD are achieved as the student moves away from the teacher, rather than toward it.

These observations lead us to hypothesize that the performance gains of clipped OPSD arise primarily from changes in reasoning behavior as the student moves away from the privileged teacher, rather than from the transfer of privileged knowledge. Access to privileged information allows the teacher to make more confident predictions with less need for extensive reasoning~\citep{baumann2026repulsive}. The teacher distribution may therefore act as a thinking switch: moving toward it suppresses thinking, whereas moving away from it promotes more extensive reasoning, as illustrated in Fig.~\ref{fig:student-teacher-thinking}.

To validate this hypothesis, we introduce \textbf{Teacher Reversal Alignment (TRA)}, which constructs a target distribution proportional to $p^2/q$, where $p$ and $q$ denote the student and privileged teacher distributions, respectively. By explicitly reversing the direction of distributional alignment, TRA drives the student away from the privileged teacher. As shown in Fig.~\ref{fig:tra-aime24}, TRA achieves the best reasoning performance among all compared methods across Qwen3-1.7B, Qwen3-4B, and Qwen3-8B, reaching AIME24 Avg@16 scores of 41.0\%, 67.1\%, and 73.8\%, respectively. These results represent improvements of 27.2, 44.0, and 47.3 percentage points over the base models, and 22.1, 24.3, and 15.6 percentage points over the strongest competing methods. These gains are accompanied by substantial increases in reasoning length, reaching 5.58, 3.75, and 3.80 times those of the base models, respectively. These results demonstrate that substantial reasoning gains can be achieved by moving away from privileged teachers rather than aligning with them. We further validate this hypothesis through extensive experiments across diverse settings, including irrelevant privileged solutions, answer-only privileged information, and stronger privileged teachers. All these experiments converge on a unified conclusion: \textbf{privileged teachers were never about teaching knowledge; they were thinking switches all along.}

\section{Background}

\paragraph{On-Policy Self-Distillation.}
On-policy distillation (OPD) provides dense, token-level supervision along student-generated trajectories, alleviating the distribution mismatch associated with off-policy distillation~\citep{agarwal2024policy}. Unlike conventional OPD, which typically relies on a separate and often stronger teacher model, On-Policy Self-Distillation (OPSD) employs the same model in two distinct contextual roles. Specifically, the student receives only the original problem, whereas the privileged teacher additionally accesses information unavailable to the student, typically a verified solution, and provides distributional supervision along student-generated trajectories~\citep{zhao2026self}. This formulation incorporates privileged information into on-policy training without requiring an external teacher model. Subsequent studies have extended this paradigm to various settings, including agentic tasks~\citep{penaloza2026privileged}, reference-signal purification~\citep{shen2026purified}, privileged-context selection~\citep{tian2026should}, cross-problem reference substitution~\citep{ichihara2026privileged}, verified rollout supervision~\citep{hossain2026overcoming}, self-generated problem briefs~\citep{xu2026b} and curricula~\citep{sick2026policy}, as well as self-consistency-based pseudo-solutions~\citep{li2026policy}. Related approaches further incorporate environment feedback~\citep{hubotter2026reinforcement} and reward-conditioned self-revision~\citep{he2026self}. More recent studies have begun to examine the role of privileged information itself, particularly its contribution to reasoning performance and its influence on long-chain-of-thought (CoT) reasoning~\citep{kaur2026rethinking}.

\paragraph{Reasoning Enhancement.}
Chain-of-thought (CoT) prompting encourages language models to generate intermediate reasoning steps before arriving at final answers~\citep{wei2022chain, kojima2022large}. Beyond explicit reasoning elicitation, self-consistency and tree-of-thoughts enhance problem-solving capabilities by exploring and aggregating multiple reasoning paths~\citep{wang2022self, yao2023tree}. Subsequent studies have demonstrated that scaling inference-time computation can further improve reasoning performance~\citep{snell2025scaling}, with growing attention to optimal reasoning-length allocation~\citep{yang2026towards}, reinforcement learning for extended reasoning~\citep{guo2025deepseek}, and budget-aware reasoning evaluation~\citep{wang2024reasoning, sun2025empirical}. Building on these findings, more recent work has explored explicit control over reasoning effort through test-time budget forcing~\citep{muennighoff2025s1}, reinforcement-learning-based length control~\citep{aggarwal2025l1}, inference-aware optimization~\citep{yu2025think}, and adaptive reasoning termination~\citep{wu2025thought, han2026your}. Collectively, these studies establish reasoning effort as a central dimension of inference-time computation and highlight the potential of allocating additional computation to improve reasoning performance. However, the relationship between privileged supervision and reasoning effort remains largely unexplored.

\begin{figure*}[t]
    \centering
    \begin{subfigure}[t]{0.24\textwidth}
        \centering
        \includegraphics[width=\linewidth]{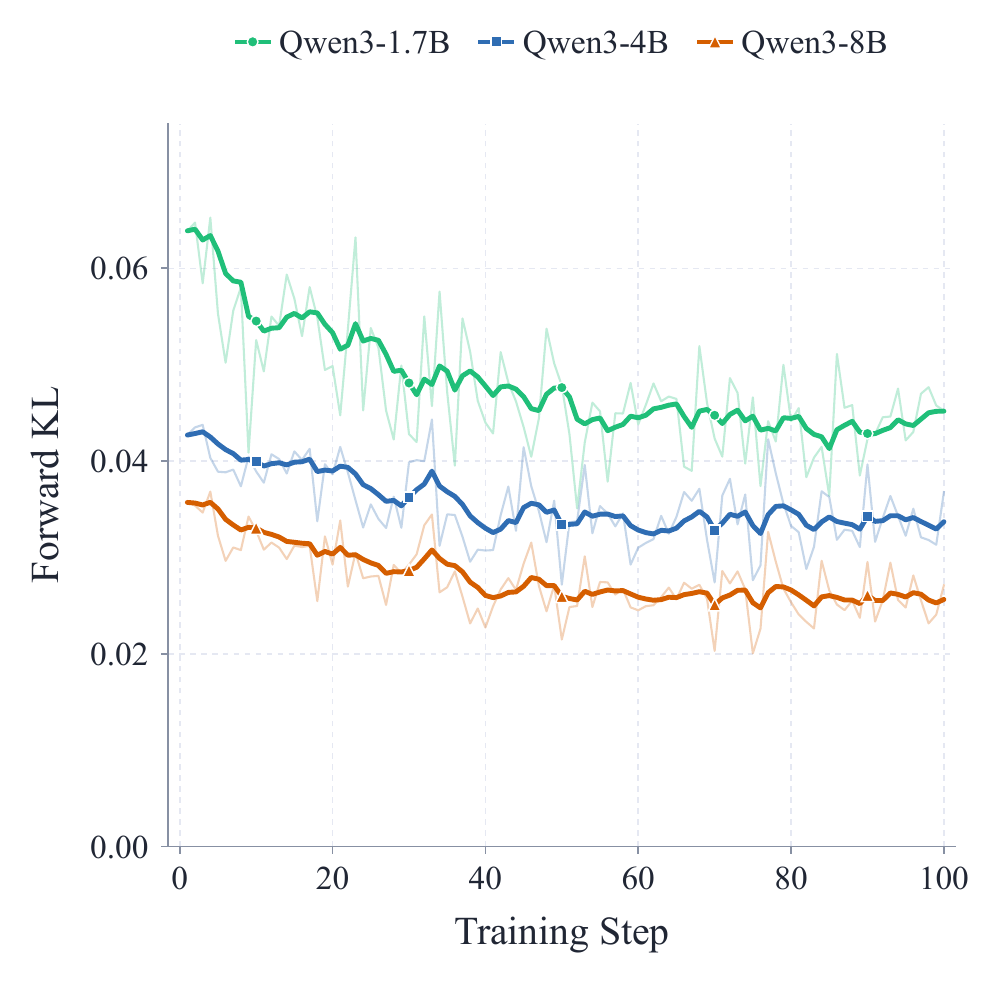}
        \caption{Student--teacher KL.}
    \end{subfigure}\hfill
    \begin{subfigure}[t]{0.24\textwidth}
        \centering
        \includegraphics[width=\linewidth]{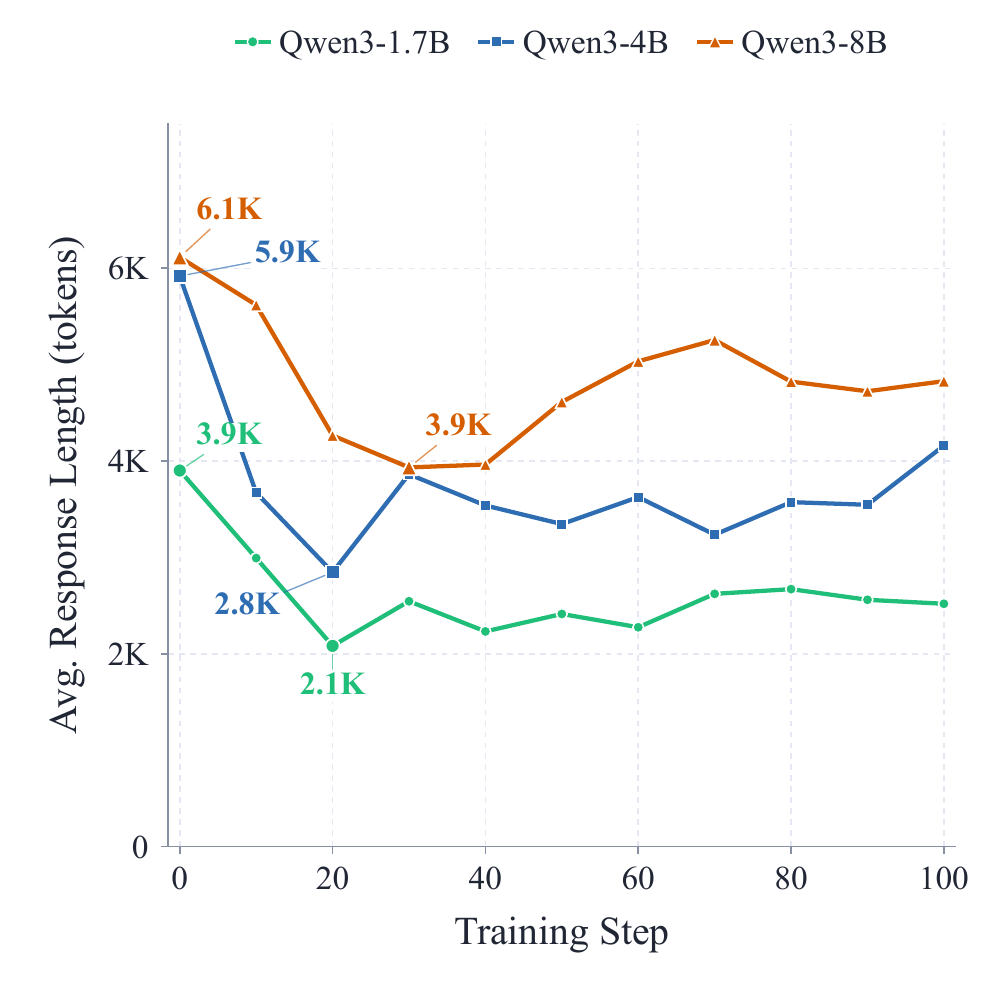}
        \caption{Reasoning length.}
    \end{subfigure}\hfill
    \begin{subfigure}[t]{0.24\textwidth}
        \centering
        \includegraphics[width=\linewidth]{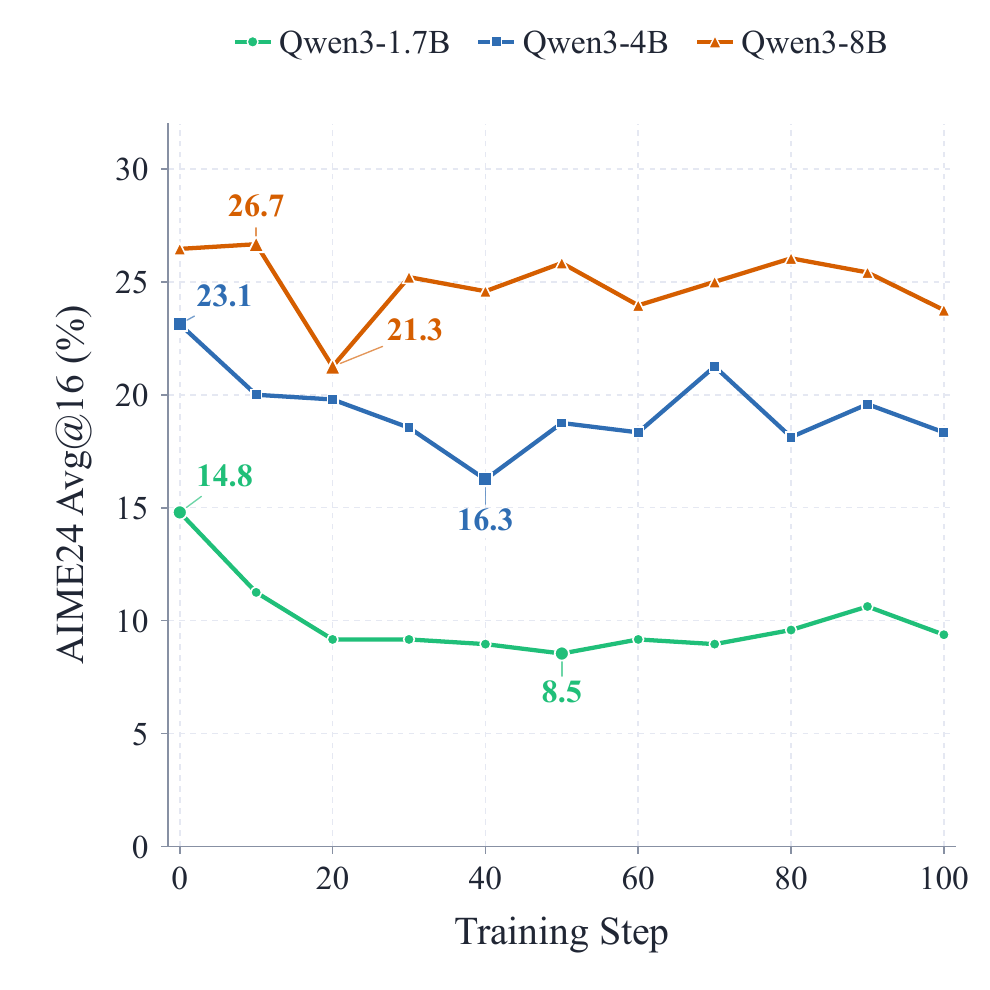}
        \caption{Avg@16.}
    \end{subfigure}\hfill
    \begin{subfigure}[t]{0.24\textwidth}
        \centering
        \includegraphics[width=\linewidth]{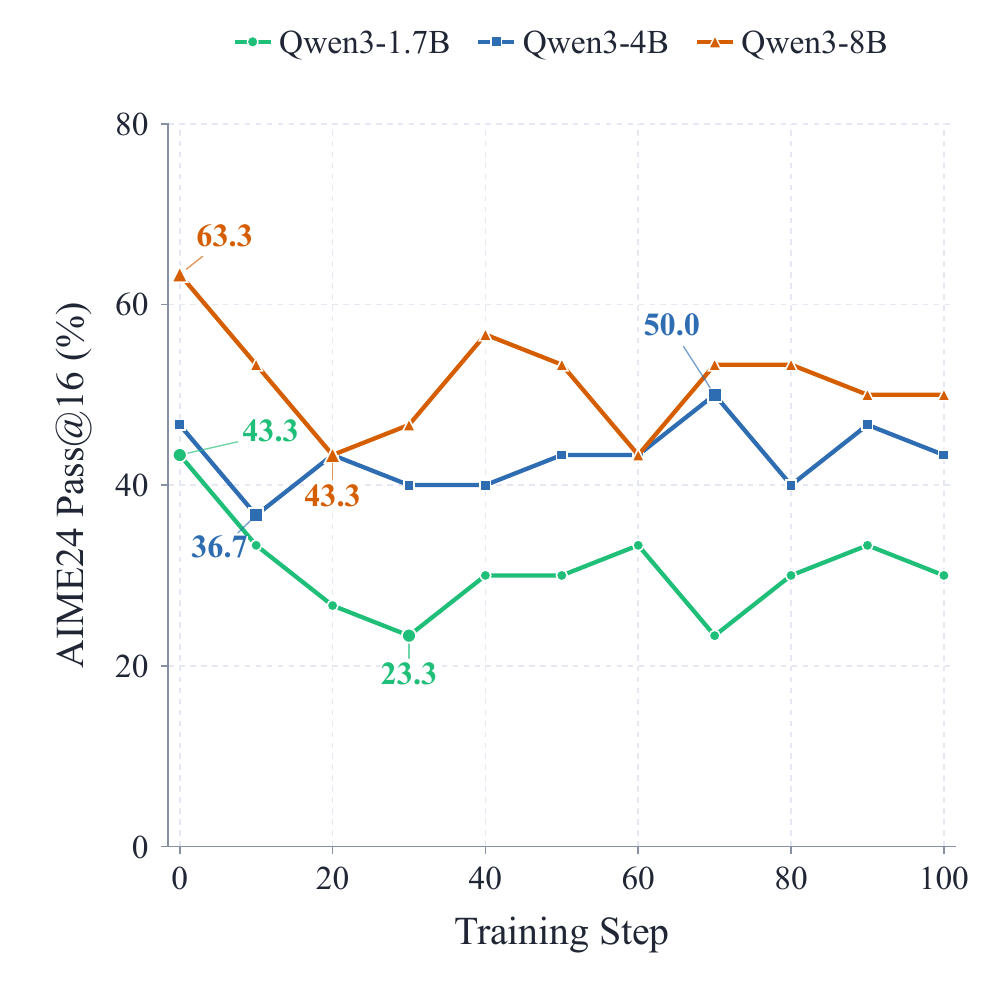}
        \caption{Pass@16.}
    \end{subfigure}
    \caption{Training dynamics of direct forward KL.}
    \label{fig:direct-fkl}
\end{figure*}

\begin{figure*}[t]
    \centering
    \begin{subfigure}[t]{0.24\textwidth}
        \centering
        \includegraphics[width=\linewidth]{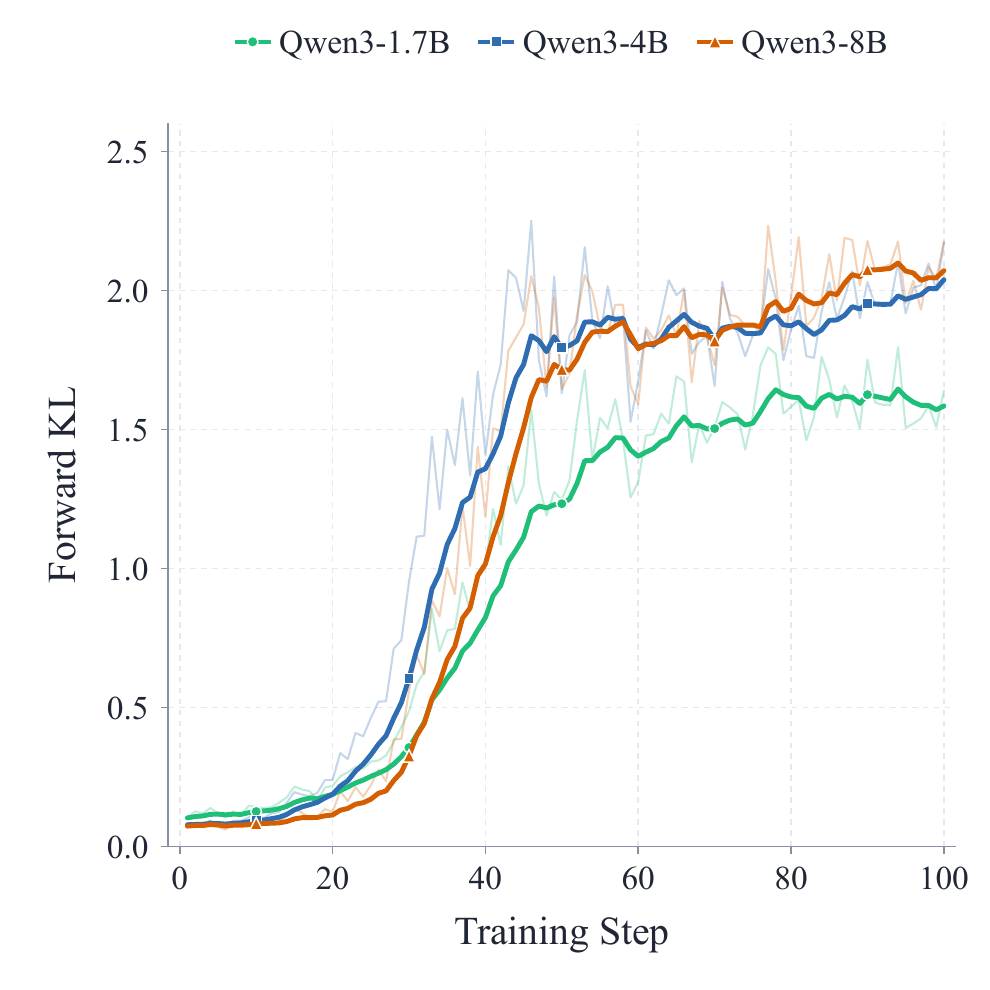}
        \caption{Student--teacher KL.}
    \end{subfigure}\hfill
    \begin{subfigure}[t]{0.24\textwidth}
        \centering
        \includegraphics[width=\linewidth]{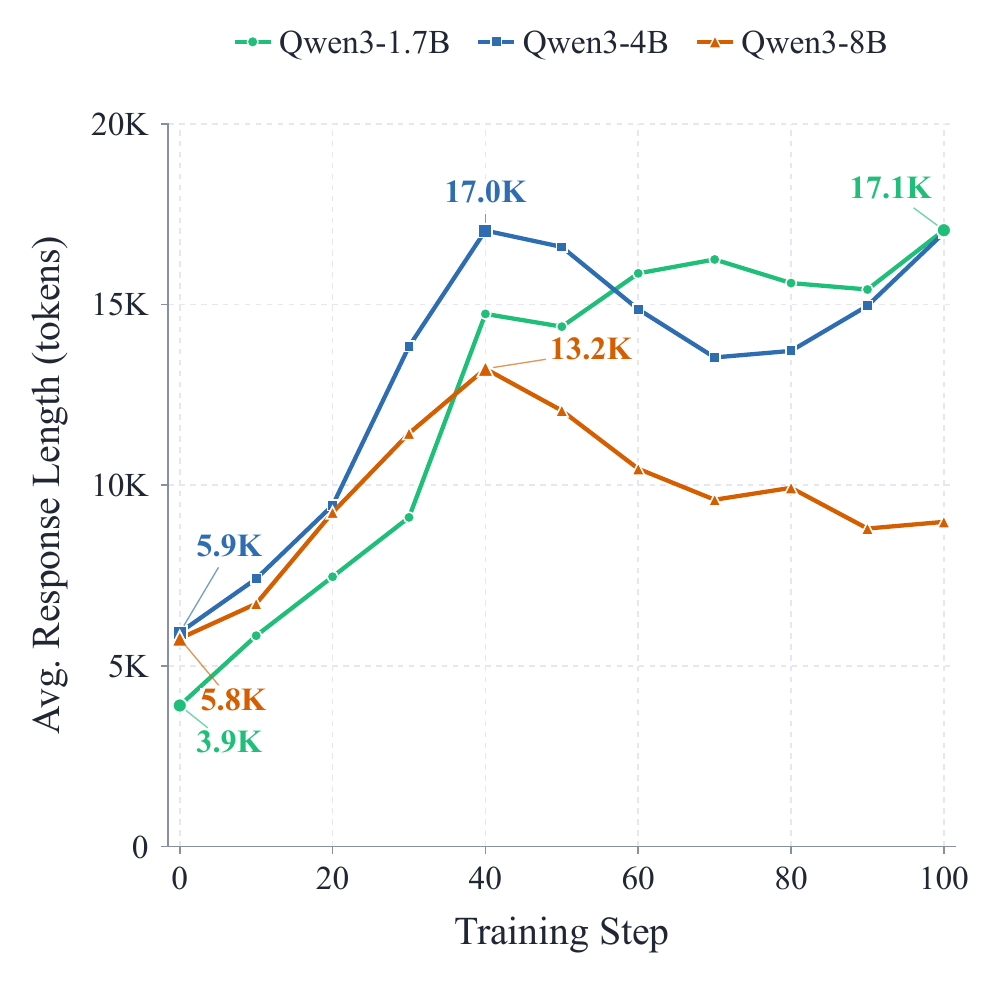}
        \caption{Reasoning length.}
    \end{subfigure}\hfill
    \begin{subfigure}[t]{0.24\textwidth}
        \centering
        \includegraphics[width=\linewidth]{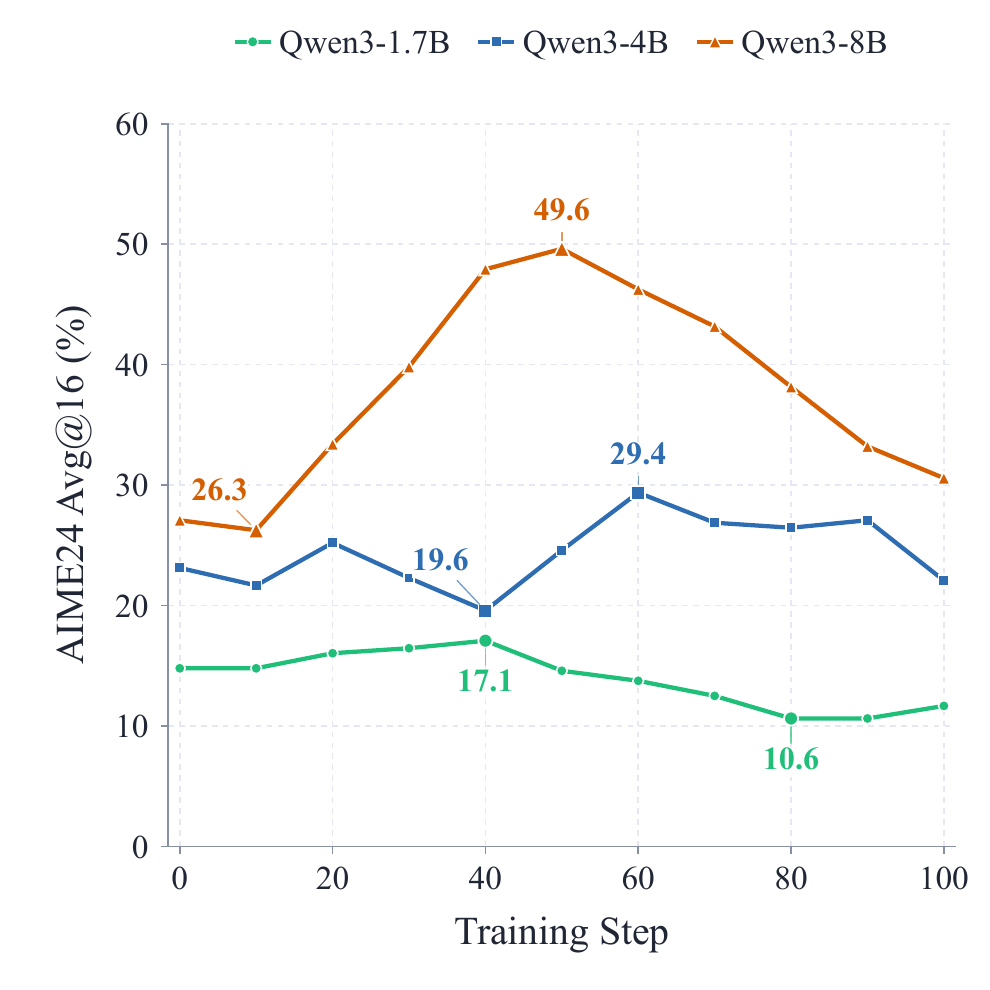}
        \caption{Avg@16.}
    \end{subfigure}\hfill
    \begin{subfigure}[t]{0.24\textwidth}
        \centering
        \includegraphics[width=\linewidth]{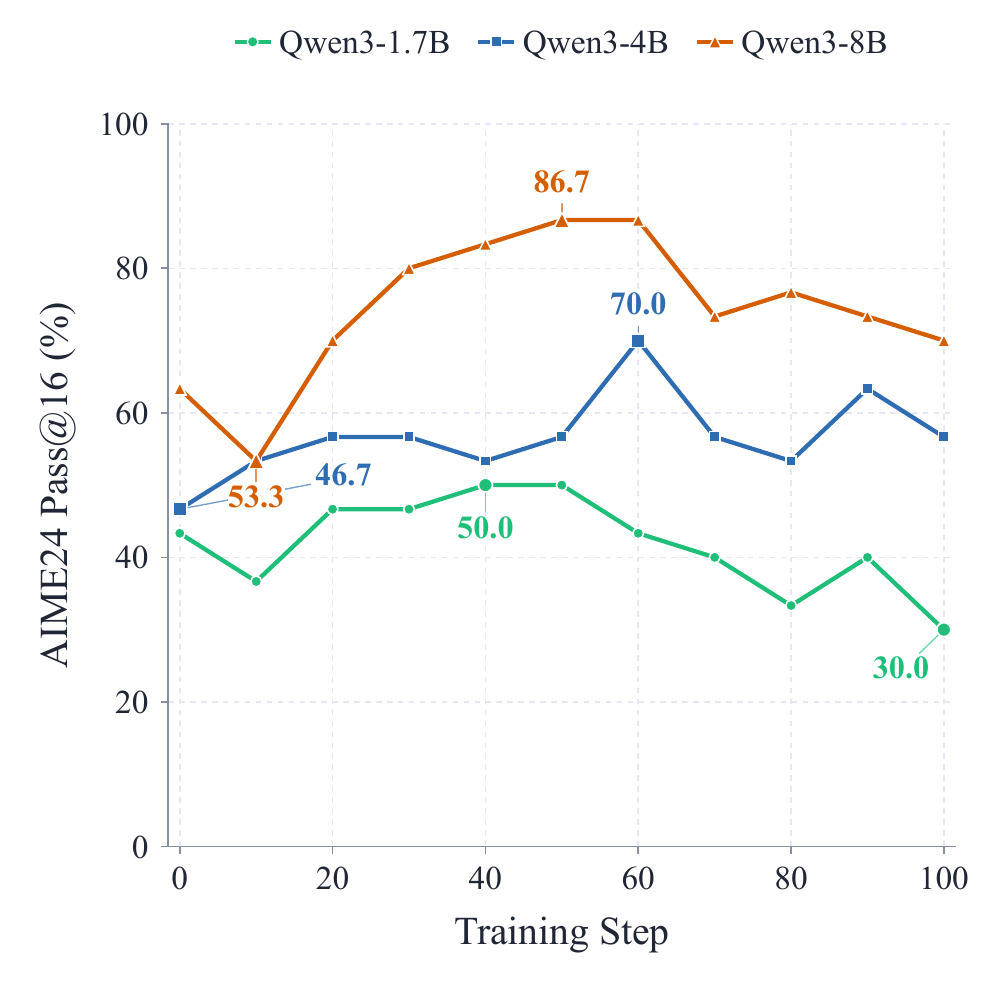}
        \caption{Pass@16.}
    \end{subfigure}
    \caption{Training dynamics of Clip-OPSD.}
    \label{fig:clip-opsd}
\end{figure*}

\section{Experimental Setup}

\subsection{Models and Tasks}

We conduct experiments using three models from the Qwen3 family: Qwen3-1.7B, Qwen3-4B, and Qwen3-8B~\citep{yang2025qwen3}. Crucially, \textbf{\uline{both the student and the privileged teacher operate in non-thinking mode, and all training and evaluation are conducted exclusively in this mode.}} The explicit thinking mode is never activated at any stage of our experiments. For training, we use the mathematical subset of OpenThoughts-114k~\citep{guha2026openthoughts}. Each example consists of a mathematical problem paired with an explicit reasoning trace and its corresponding answer, providing a consistent source of privileged information across all three models.

\subsection{Training Configuration}

All models are fine-tuned using LoRA with a rank of 64 and an alpha of 128, with a learning rate of $5\times10^{-6}$. All training is conducted in BF16 precision, with gradient norm clipping set to 0.1. For all methods, the privileged teacher remains fixed throughout training and is not updated alongside the student. Each training batch consists of 32 questions, with 8 responses sampled per question. We train each model for 100 optimization steps, using a maximum generation length of 8,192 tokens. Response generation uses a temperature of 1.0 and a top-$p$ value of 1.0. We use a fixed training seed of 42 for all experiments.

\subsection{Evaluation Protocol}

We evaluate on AIME24, consisting of 30 problems, before training and every 10 optimization steps. For each problem, we sample 16 responses using a temperature of 0.6 and top-$p$ of 0.95, with a maximum generation length of 32,768 tokens. All evaluations use question-only prompts without privileged solutions, with thinking mode disabled. We report Avg@16, Pass@16, and average response length. Avg@16 measures the mean accuracy of the 16 sampled responses across all problems, while Pass@16 measures the proportion of problems for which at least one of the 16 responses is correct. We use a fixed evaluation seed of 42.

\section{Revisiting On-Policy Self-Distillation}

\subsection{Formulation of OPSD}

Given a problem $x$ and its verified solution $s$, On-Policy Self-Distillation (OPSD) uses the same model as both student and privileged teacher. The student generates a trajectory $y\sim p_\theta(\cdot\mid x)$, while the teacher additionally accesses the verified solution $s$. At each token position $t$, their next-token distributions are:
\begin{equation}
    p_\theta^t = p_\theta(\cdot\mid x,y_{<t}),
    \qquad
    q_\theta^t = p_\theta(\cdot\mid x,s,y_{<t}).
\end{equation}
OPSD aims to minimize the forward KL divergence between the privileged teacher and student distributions over student-generated trajectories:
\begin{equation}
    \mathcal{L}_{\mathrm{OPSD}}(\theta)
    =
    \mathbb{E}_{\substack{(x,s)\sim\mathcal{D}\\
    y\sim p_\theta(\cdot\mid x)}}
    \left[
    \frac{1}{T}\sum_{t=1}^{T}
    D_{\mathrm{KL}}
    \left(
    \operatorname{sg}(q_\theta^t)
    \,\|\,p_\theta^t
    \right)
    \right],
    \label{eq:opsd}
\end{equation}
where $T$ is the trajectory length and $\operatorname{sg}(\cdot)$ denotes stop-gradient. During each optimization step, the sampled trajectories are treated as fixed.

\subsection{Direct Forward-KL Alignment}

We first investigate standard OPSD by directly optimizing the forward-KL objective defined in Eq.~\eqref{eq:opsd}, using verified solutions as privileged information.

As shown in Fig.~\ref{fig:direct-fkl}, direct forward-KL alignment consistently reduces the student--teacher KL divergence across all three models, indicating progressively closer alignment with privileged teachers. Meanwhile, reasoning lengths and both AIME24 performance metrics generally decline throughout training. For example, Qwen3-1.7B exhibits a reduction in forward KL from approximately 0.064 to 0.045, while its average response length decreases from 3.9K to 2.5K tokens. Correspondingly, Avg@16 drops from 14.8\% to approximately 9.4\%, and Pass@16 decreases from 43.3\% to 30.0\%. The concurrent decline in Pass@16 further indicates that the degradation in reasoning performance persists even when allowing 16 attempts per problem. Similar trends are observed in Qwen3-4B and Qwen3-8B, demonstrating that direct alignment toward privileged teachers ultimately reduces reasoning length and degrades task performance across all three models.

\begin{keynote}{Finding 1: Less Thinking}
\textbf{Privileged alignment suppresses reasoning.}

Moving toward the teacher reduces thinking and degrades performance.
\end{keynote}

\subsection{Clipped Forward-KL Alignment}

OPSD adopts a clipped forward-KL objective, which we refer to as \textbf{Clip-OPSD}~\citep{zhao2026self}. The underlying intuition is that privileged teachers, having access to verified solutions, produce more confident predictions and assign lower probabilities to exploratory tokens such as ``wait''. Aligning the student with these predictions is therefore expected to suppress exploratory reasoning and encourage shorter responses. To stabilize distributional alignment, OPSD applies pointwise clipping to individual vocabulary-level KL contributions. Specifically, the objective is defined as
\begin{equation}
\begin{aligned}
\mathcal{L}_{\mathrm{Clip\text{-}OPSD}}(\theta)
&=
\mathbb{E}_{\substack{(x,s)\sim\mathcal{D}\\
y\sim p_\theta(\cdot\mid x)}}
\Bigg[
\frac{1}{T}\sum_{t=1}^{T}
\\
&\quad \sum_{v\in\mathcal{V}}
\min\left(
q_\theta^t(v)\log\frac{q_\theta^t(v)}
{p_\theta^t(v)},
\epsilon
\right)
\Bigg],
\end{aligned}
\label{eq:clip-opsd}
\end{equation}
where $\epsilon=10^{-7}$ is the clipping threshold. Clipping is applied independently to each vocabulary-level KL contribution before summation.

As shown in Fig.~\ref{fig:clip-opsd}, Clip-OPSD substantially increases the student--teacher KL divergence across all three models, indicating that students progressively move away from their privileged teachers despite the nominal forward-KL objective. This divergence is accompanied by pronounced increases in reasoning length and performance gains, although the magnitude of these gains varies across model scales. Notably, Qwen3-8B exhibits the most substantial improvement, with Avg@16 increasing from 26.3\% to 49.6\% and Pass@16 from 63.3\% to 86.7\%, while its average response length expands from 5.8K to approximately 13.2K tokens. These results reveal a consistent association between increasing student--teacher divergence, more extensive reasoning, and improved reasoning performance, with the strongest gains observed in Qwen3-8B.

\begin{keynote}{Finding 2: More Thinking}
\textbf{Moving away the teacher drives early gains.}

Clip-OPSD improves reasoning performance as students progressively move away from privileged teachers and expand their reasoning.
\end{keynote}

\begin{figure*}[t]
    \centering
    \begin{subfigure}[t]{0.24\textwidth}
        \centering
        \includegraphics[width=\linewidth]{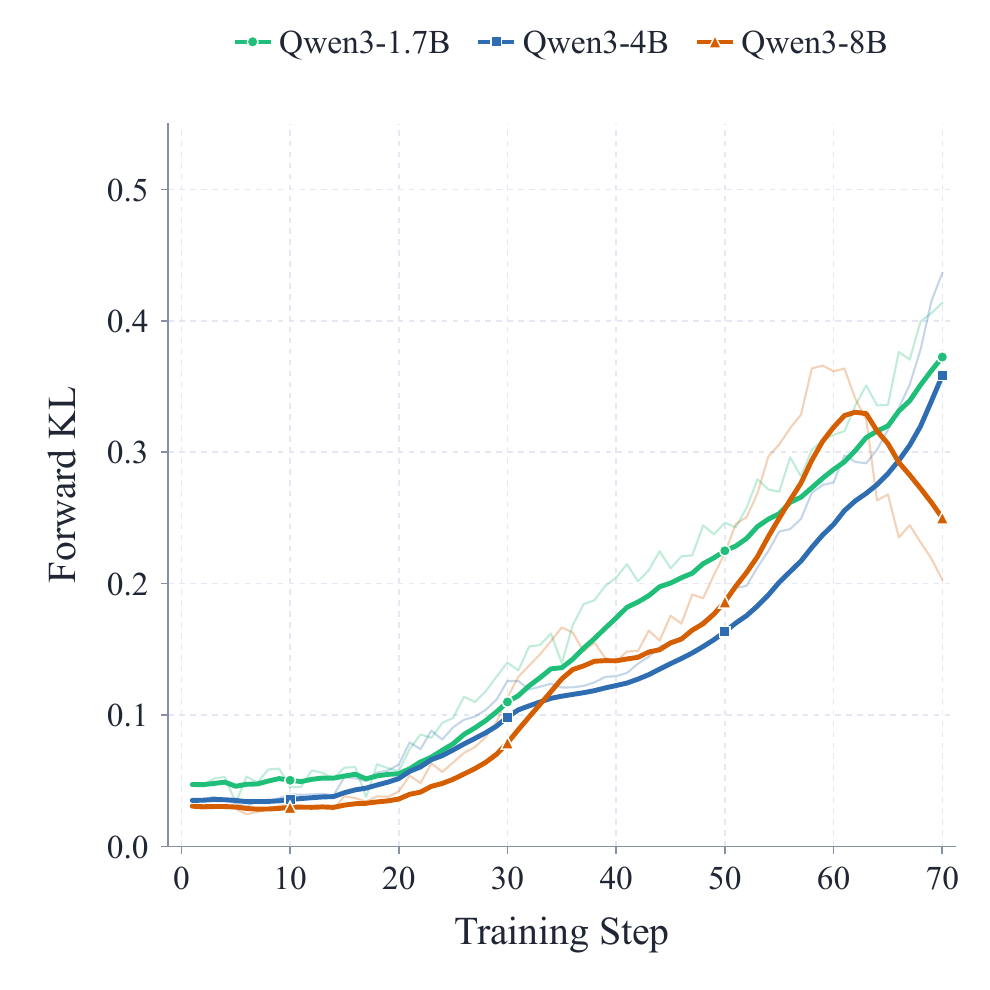}
        \caption{Student--teacher KL.}
    \end{subfigure}\hfill
    \begin{subfigure}[t]{0.24\textwidth}
        \centering
        \includegraphics[width=\linewidth]{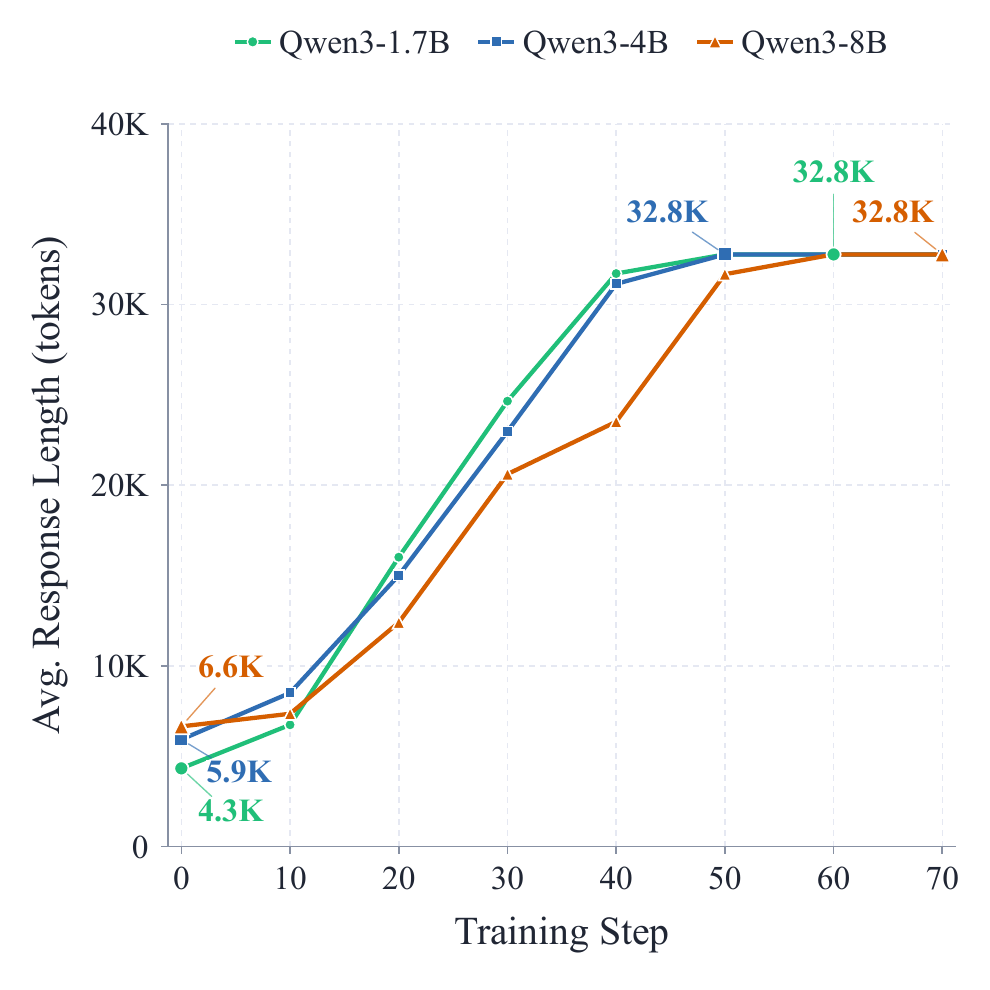}
        \caption{Reasoning length.}
    \end{subfigure}\hfill
    \begin{subfigure}[t]{0.24\textwidth}
        \centering
        \includegraphics[width=\linewidth]{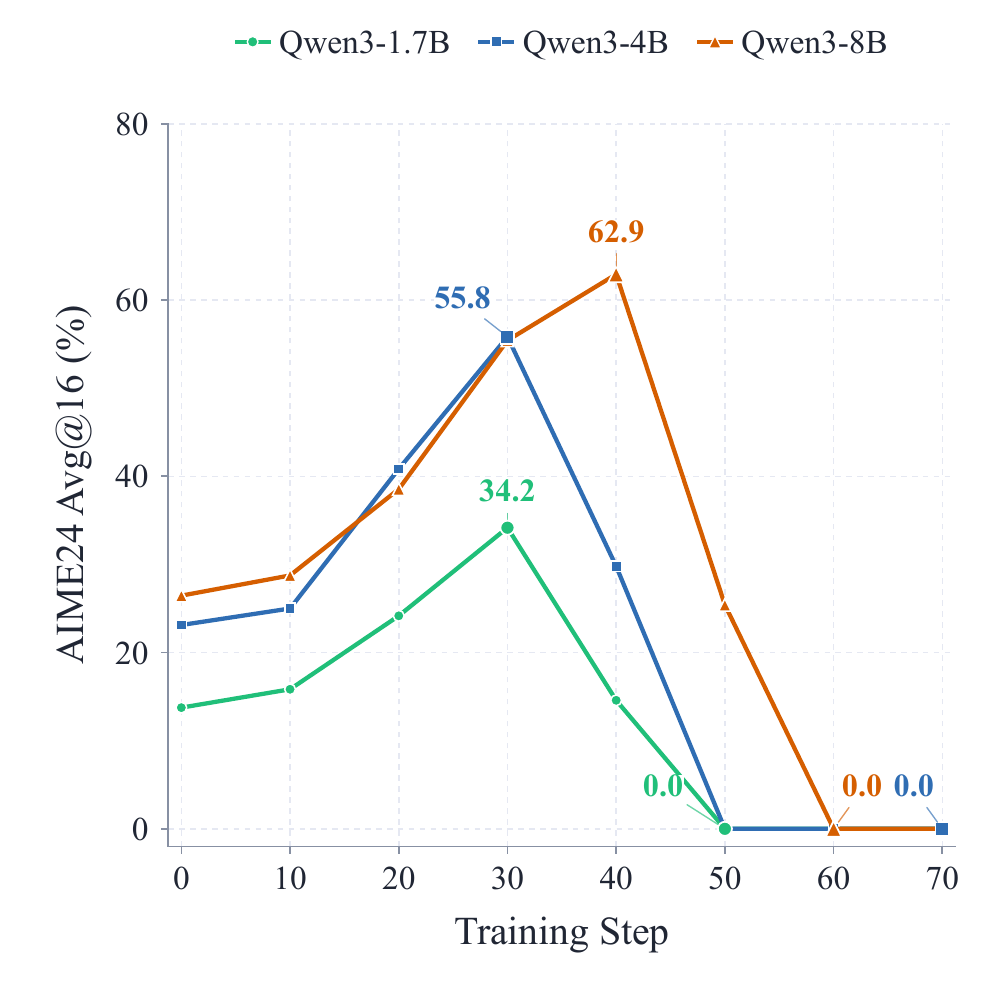}
        \caption{Avg@16.}
    \end{subfigure}\hfill
    \begin{subfigure}[t]{0.24\textwidth}
        \centering
        \includegraphics[width=\linewidth]{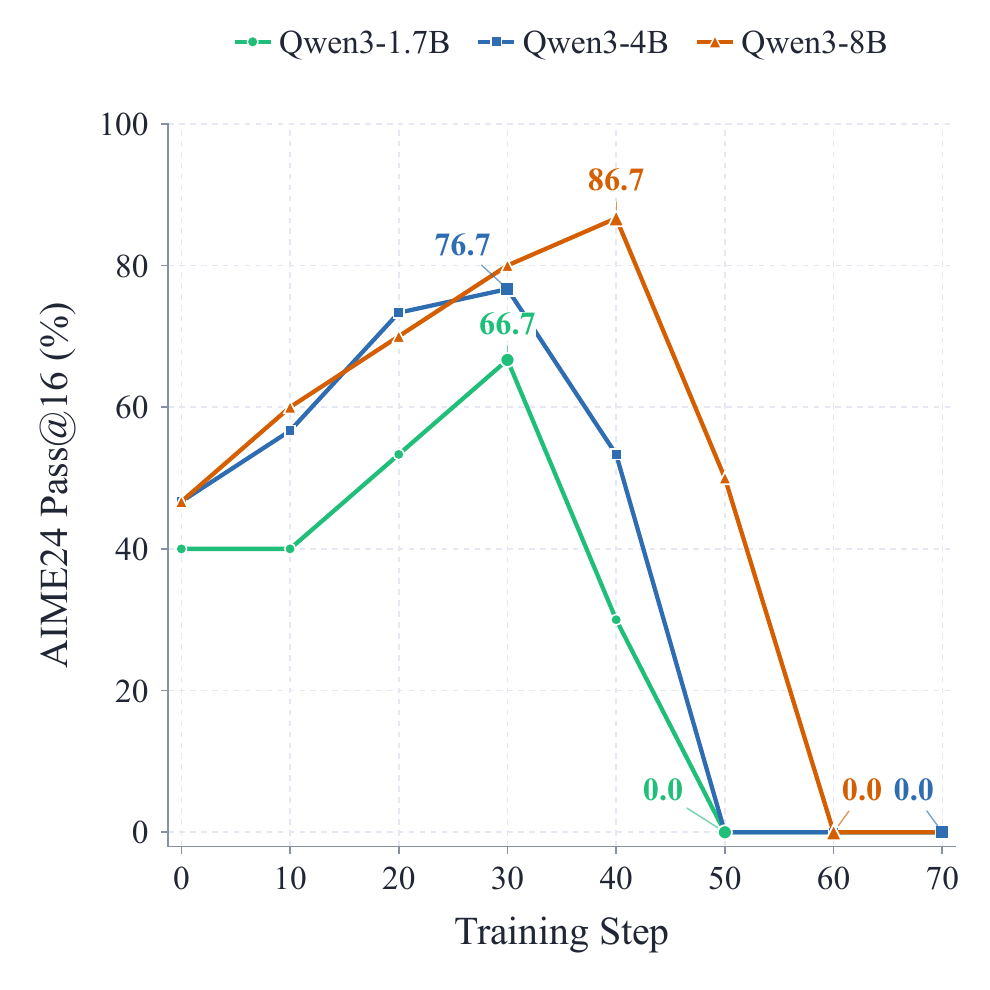}
        \caption{Pass@16.}
    \end{subfigure}
    \caption{Training dynamics of TRA ($n=1$).}
    \label{fig:tra-n1}
\end{figure*}

\begin{figure*}[t]
    \centering
    \begin{subfigure}[t]{0.24\textwidth}
        \centering
        \includegraphics[width=\linewidth]{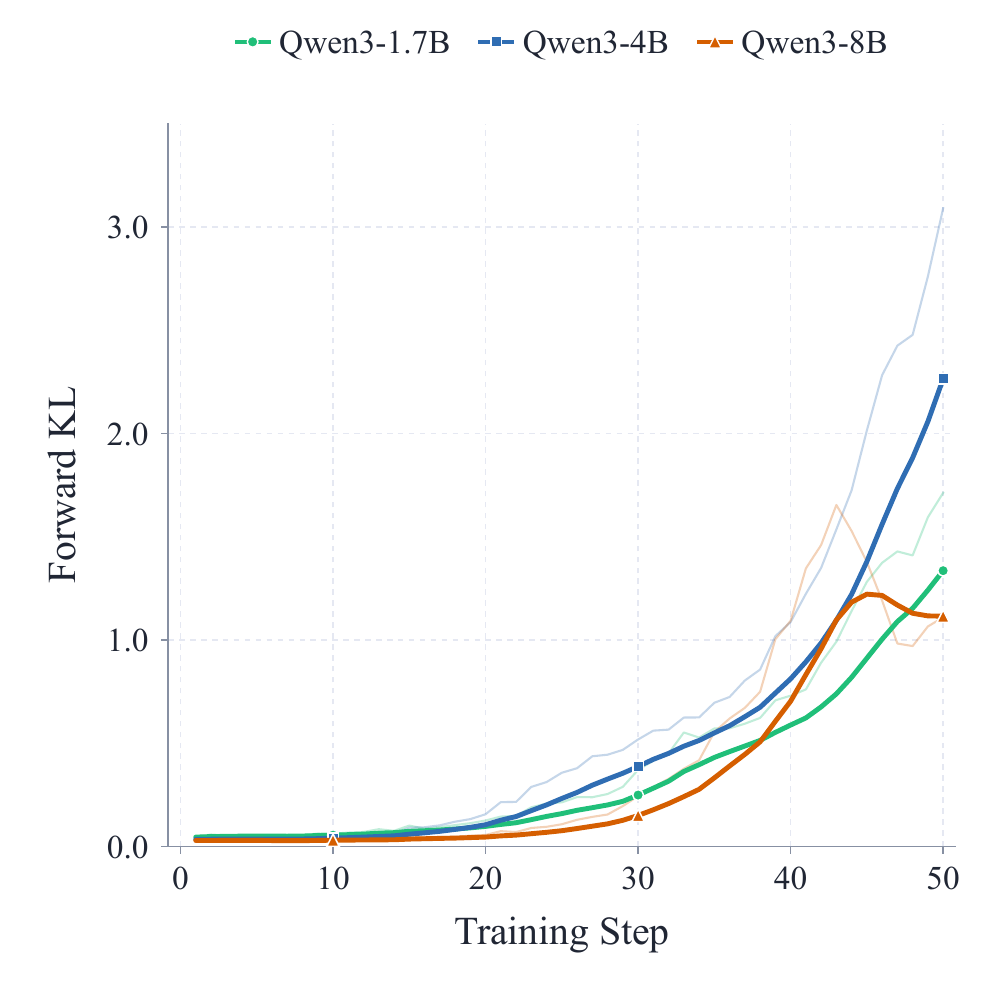}
        \caption{Student--teacher KL.}
    \end{subfigure}\hfill
    \begin{subfigure}[t]{0.24\textwidth}
        \centering
        \includegraphics[width=\linewidth]{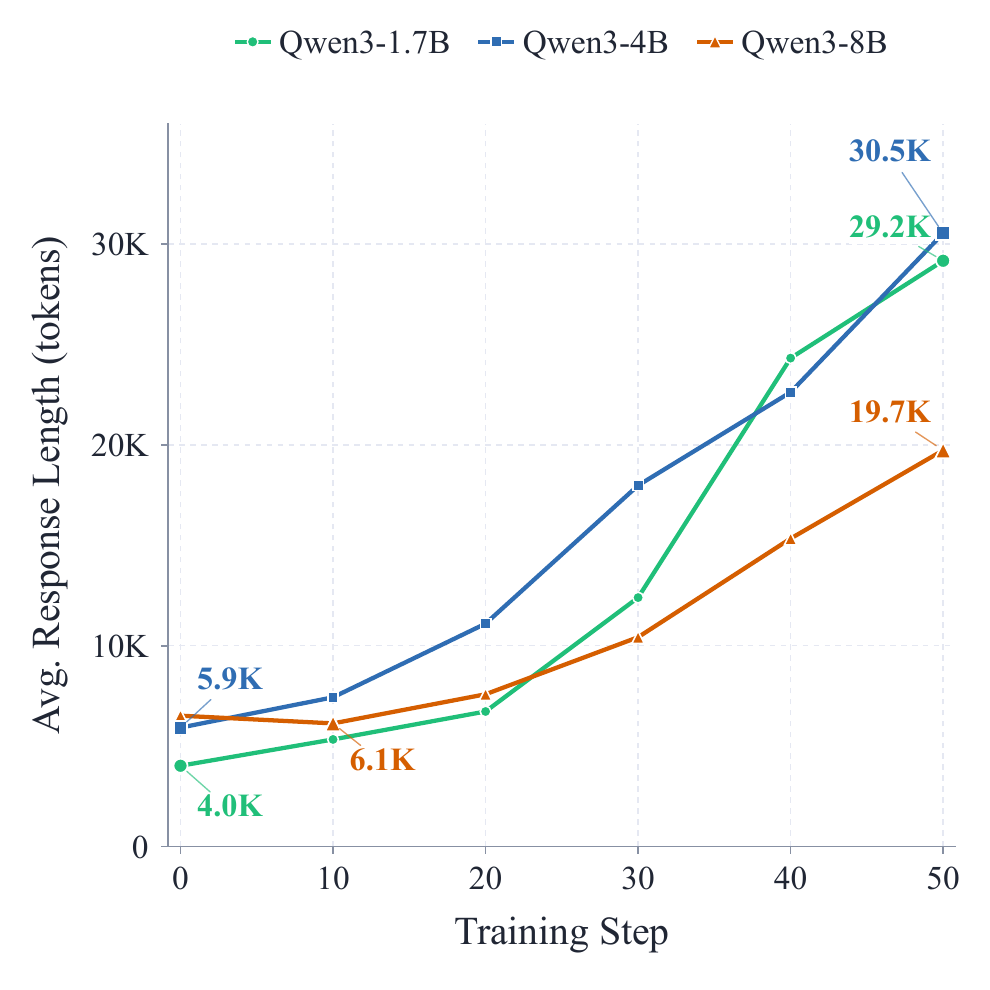}
        \caption{Reasoning length.}
    \end{subfigure}\hfill
    \begin{subfigure}[t]{0.24\textwidth}
        \centering
        \includegraphics[width=\linewidth]{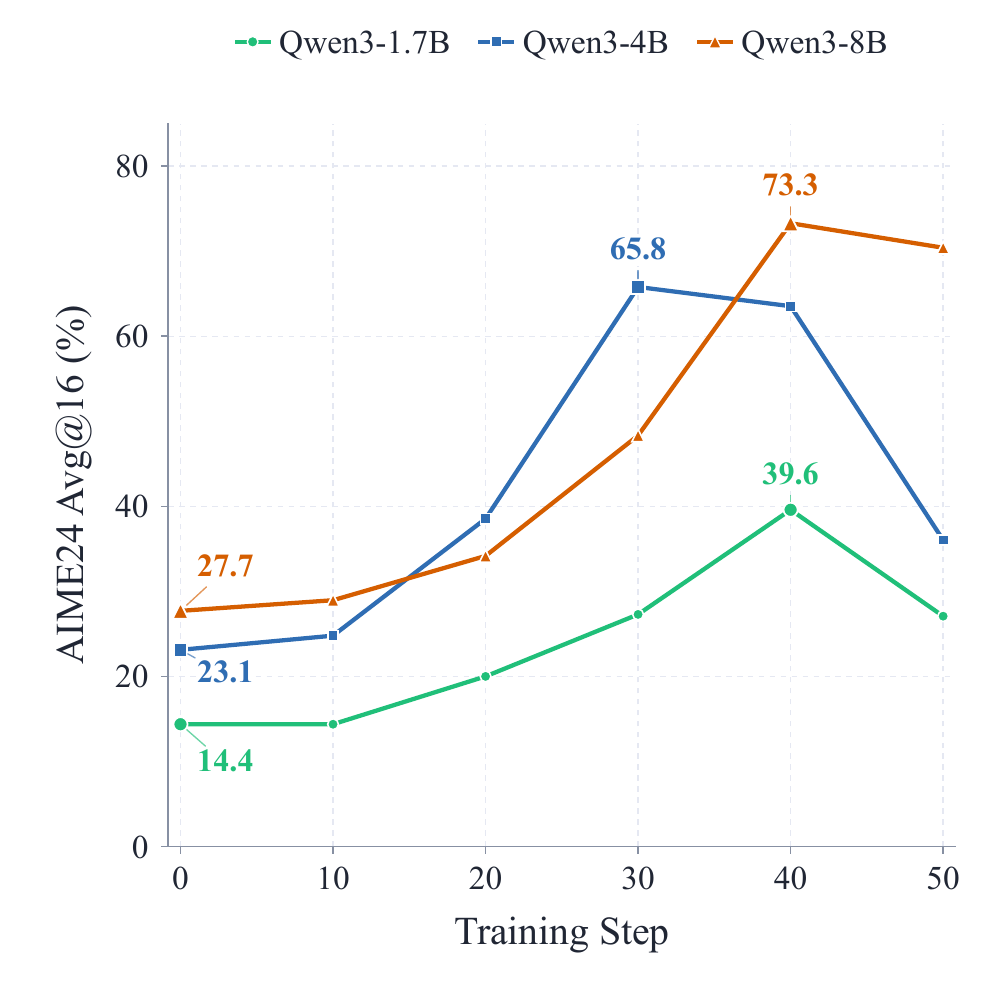}
        \caption{Avg@16.}
    \end{subfigure}\hfill
    \begin{subfigure}[t]{0.24\textwidth}
        \centering
        \includegraphics[width=\linewidth]{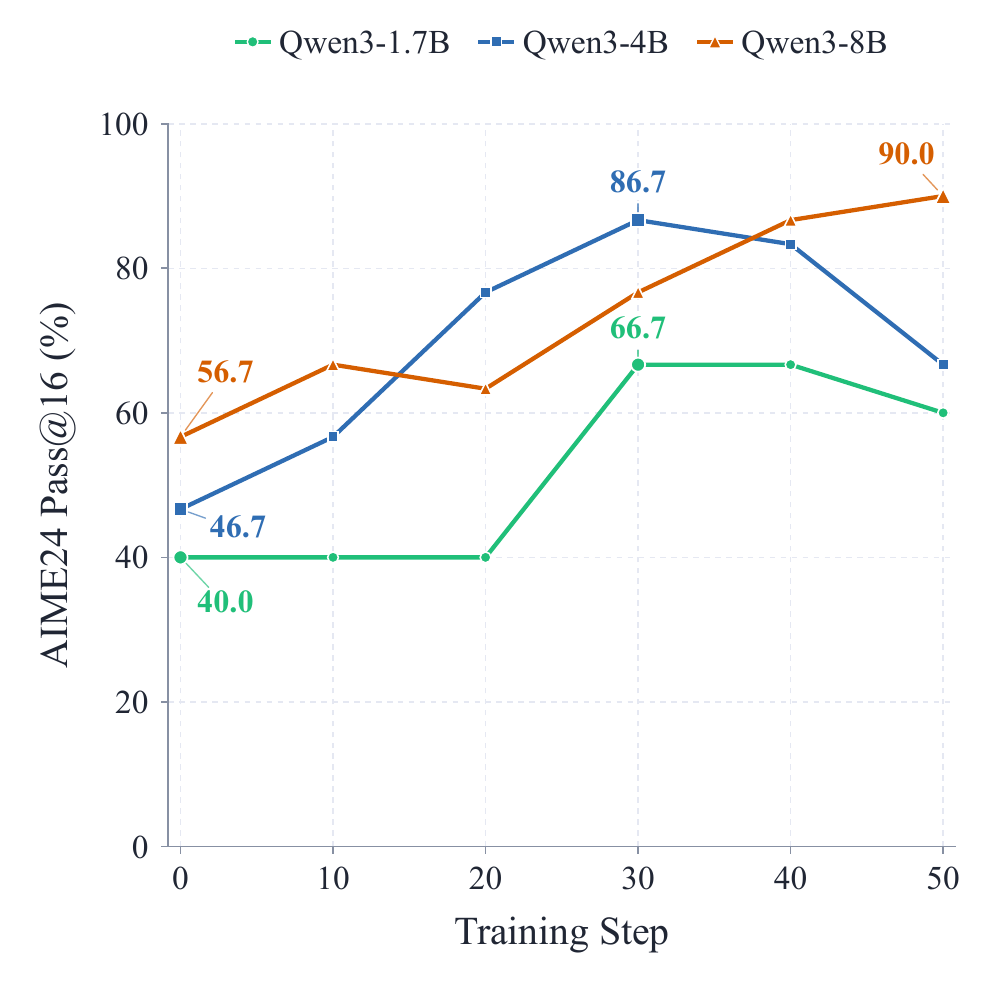}
        \caption{Pass@16.}
    \end{subfigure}
    \caption{Training dynamics of TRA ($n=0.1$).}
    \label{fig:tra-n0p1}
\end{figure*}

\subsection{Understanding the Effective Optimization Direction}

Although Clip-OPSD is formulated as forward-KL alignment, its clipping operation fundamentally changes the effective target distribution. We derive an equivalent distributional objective to characterize its actual optimization direction. For simplicity, we consider a single token position and omit the position index. Let $p$ and $q$ denote the student and fixed privileged teacher distributions over vocabulary $\mathcal{V}$, respectively. Define the individual KL contribution as
$c_i=q_i\log(q_i/p_i)$, and partition the vocabulary into clipped and unclipped sets:
\begin{equation}
    \begin{aligned}
        \mathcal{C}
        &=\{i\in\mathcal{V}:c_i>\epsilon\},\\
        \mathcal{U}
        &=\mathcal{V}\setminus\mathcal{C},
    \end{aligned}
\end{equation}
where $\epsilon=10^{-7}$. We further denote the clipped teacher probability mass by
\begin{equation}
    Q_{\mathcal{C}}=\sum_{i\in\mathcal{C}}q_i,
    \qquad
    Q_{\mathcal{U}}=1-Q_{\mathcal{C}}.
\end{equation}

\textbf{Equivalent Target Distribution.}
Under a fixed clipping partition, only unclipped terms contribute to the student gradient. Let $z_j$ denote the student logit for token $j$. The gradient of Clip-OPSD is
\begin{equation}
    \frac{\partial\mathcal{L}_{\mathrm{clip}}}{\partial z_j}
    =
    Q_{\mathcal{U}}p_j
    -
    q_j\mathbb{I}(j\in\mathcal{U}).
    \label{eq:clip-gradient}
\end{equation}

This gradient can be equivalently expressed as forward-KL alignment toward a reallocated target distribution:
\begin{equation}
    \boxed{
    r_i
    =
    q_i\mathbb{I}(i\in\mathcal{U})
    +
    Q_{\mathcal{C}}p_i
    }
    \label{eq:clip-target}
\end{equation}
Specifically, the unclipped tokens retain their original teacher probabilities and receive additional probability mass proportional to the student distribution, whereas clipped tokens receive only the redistributed student probability mass. Importantly, $r$ is a valid probability distribution:
\begin{equation}
    \sum_{i\in\mathcal{V}}r_i
    =
    \sum_{i\in\mathcal{U}}q_i
    +
    Q_{\mathcal{C}}\sum_{i\in\mathcal{V}}p_i
    =1.
\end{equation}

\textbf{Gradient Equivalence.}
We now show that optimizing the forward KL toward this reallocated distribution produces exactly the same student-logit gradient as Clip-OPSD. Consider
\begin{equation}
    \mathcal{L}_{\mathrm{equiv}}
    =
    D_{\mathrm{KL}}\left(
    \operatorname{sg}(r)\,\|\,p
    \right),
\end{equation}
where $\operatorname{sg}(\cdot)$ stops gradients through the dynamically constructed target $r$. Its gradient with respect to student logit $z_j$ is
\begin{equation}
    \begin{aligned}
    \frac{\partial\mathcal{L}_{\mathrm{equiv}}}{\partial z_j}
    &=p_j-r_j\\
    &=(1-Q_{\mathcal{C}})p_j
      -q_j\mathbb{I}(j\in\mathcal{U})\\
    &=Q_{\mathcal{U}}p_j
      -q_j\mathbb{I}(j\in\mathcal{U}).
    \end{aligned}
\end{equation}
Comparing this with Eq.~\eqref{eq:clip-gradient}, we obtain
\begin{equation}
    \boxed{
    \nabla_z\mathcal{L}_{\mathrm{clip}}
    =
    \nabla_z
    D_{\mathrm{KL}}\left(
    \operatorname{sg}(r)\,\|\,p
    \right)
    }
    \label{eq:clip-equivalence}
\end{equation}
Thus, Clip-OPSD is exactly gradient-equivalent to forward-KL alignment toward the reallocated distribution $r$, rather than the original privileged teacher distribution $q$.

\section{Privileged Teachers as Thinking Switches}

Based on the preceding observations, we hypothesize that the performance gains of Clip-OPSD primarily arise from moving away from privileged teachers, accompanied by increased reasoning length. We further hypothesize that privileged teachers act as \textbf{thinking switches}: moving toward them suppresses reasoning, whereas moving away encourages more extensive thinking.

\subsection{Teacher Reversal Alignment}

To explicitly investigate the effect of moving away from privileged teachers, we introduce \textbf{Teacher Reversal Alignment (TRA)}, which reverses the conventional alignment direction by constructing a target distribution that favors tokens preferred by the student over the teacher. Specifically, for each token position $t$, we define the target distribution as
\begin{equation}
    r_n^t(v)
    =
    \frac{
        p_\theta^t(v)
        \left(
        \frac{p_\theta^t(v)}{q_\theta^t(v)}
        \right)^n
    }{
        Z_n^t
    },
    \label{eq:tra}
\end{equation}
where
\begin{equation}
    Z_n^t
    =
    \sum_{v\in\mathcal V}
    p_\theta^t(v)
    \left(
    \frac{p_\theta^t(v)}{q_\theta^t(v)}
    \right)^n
\end{equation}
is the normalization constant, and $n\geq 0$ controls the strength of teacher reversal. The resulting target can also be expressed as $r_n^t(v)\propto (p_\theta^t(v))^{n+1}/(q_\theta^t(v))^n$. In particular, $n=1$ recovers the original TRA distribution $r\propto p^2/q$, while larger values of $n$ amplify the relative preference for student-favored tokens.

TRA then optimizes the student through forward-KL alignment toward this reversed target:
\begin{equation}
    \mathcal{L}_{\mathrm{TRA}}(\theta)
    =
    \mathbb{E}_{\substack{(x,s)\sim\mathcal D\\
    y\sim p_\theta(\cdot\mid x)}}
    \left[
    \frac{1}{T}\sum_{t=1}^{T}
    D_{\mathrm{KL}}
    \left(
    \operatorname{sg}(r_n^t)
    \,\|\,p_\theta^t
    \right)
    \right].
    \label{eq:tra-loss}
\end{equation}

As shown in Fig.~\ref{fig:tra-n1}, TRA with $n=1$ substantially increases the student--teacher KL divergence across all three models, indicating that students progressively move away from their privileged teachers. This divergence is accompanied by rapid reasoning expansion: starting from approximately 4.3K--6.6K tokens, the average response length approaches 32K within just 50--60 optimization steps. Notably, after only 30 steps, all three models already surpass their corresponding OPSD baselines in Avg@16. For example, Qwen3-4B improves from 23.1\% to 55.8\%, while Qwen3-8B reaches a peak of 62.9\% at 40 steps. However, as all responses eventually reach the 32K generation limit, Avg@16 drops to zero across all three models by step 60, corresponding to a 100\% truncation rate. This performance decline results from excessive reasoning and generation-budget truncation rather than mode collapse. These results demonstrate that explicitly moving students away from privileged teachers can rapidly induce more extensive reasoning and substantial performance gains.

\begin{keynote}{Finding 3: Thinking Control}
\textbf{Privileged teachers are thinking switches.}

Moving closer to or farther from privileged teachers regulates the extent of student reasoning.
\end{keynote}

\subsection{Controlling the Rate of Moving Away from the Teacher}

To investigate the effect of reversal strength on reasoning performance, we reduce $n$ from 1 to 0.1. As shown in Fig.~\ref{fig:tra-n0p1}, all three models achieve substantial performance improvements while exhibiting pronounced reasoning-length expansion. Notably, Qwen3-1.7B, Qwen3-4B, and Qwen3-8B reach peak Avg@16 scores of 39.6\%, 65.8\%, and 73.3\%, respectively, improving over $n=1$ by 5.4, 10.0, and 10.4 percentage points. Furthermore, Qwen3-8B achieves a Pass@16 of 90.0\%, demonstrating particularly strong reasoning performance. Nevertheless, reasoning lengths still reach 29.2K, 30.5K, and 19.7K tokens for the three models, respectively, within just 50 optimization steps. This suggests that $n=0.1$ may still not represent the optimal reversal strength, leaving further room to improve performance by controlling reasoning intensity under a fixed generation budget.

\begin{keynote}{Finding 4: Optimal Thinking}
\textbf{More thinking is not always better.}

Appropriate reasoning maximizes performance under a fixed inference budget.
\end{keynote}

\begin{figure*}[t]
    \centering
    \begin{subfigure}[t]{0.24\textwidth}
        \centering
        \includegraphics[width=\linewidth]{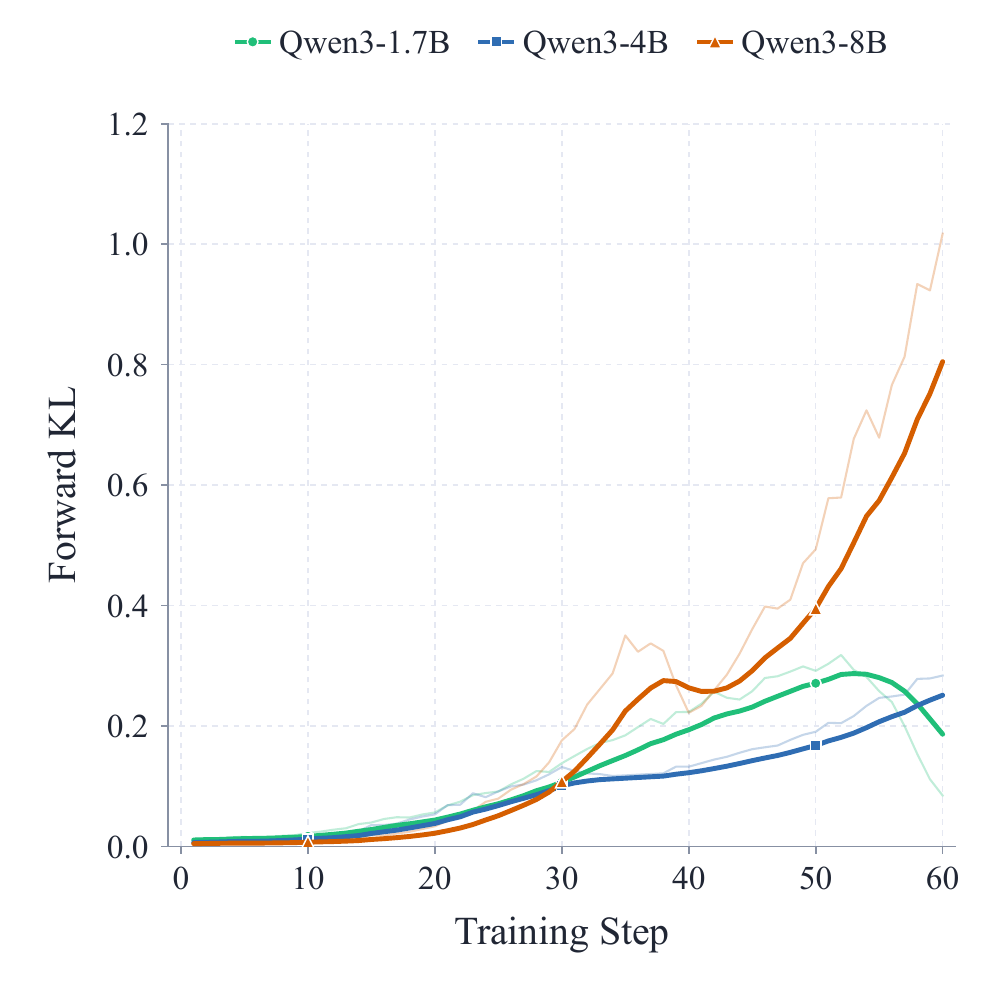}
        \caption{Student--teacher KL.}
    \end{subfigure}\hfill
    \begin{subfigure}[t]{0.24\textwidth}
        \centering
        \includegraphics[width=\linewidth]{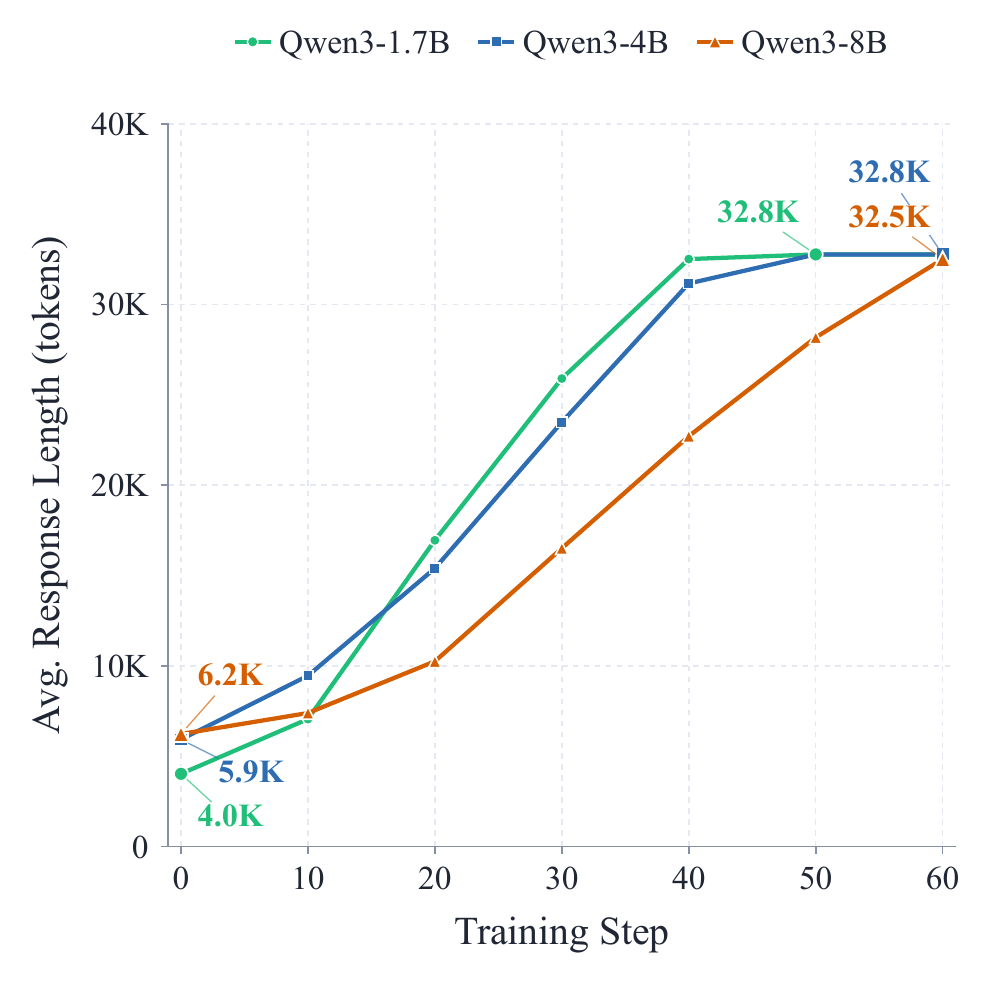}
        \caption{Reasoning length.}
    \end{subfigure}\hfill
    \begin{subfigure}[t]{0.24\textwidth}
        \centering
        \includegraphics[width=\linewidth]{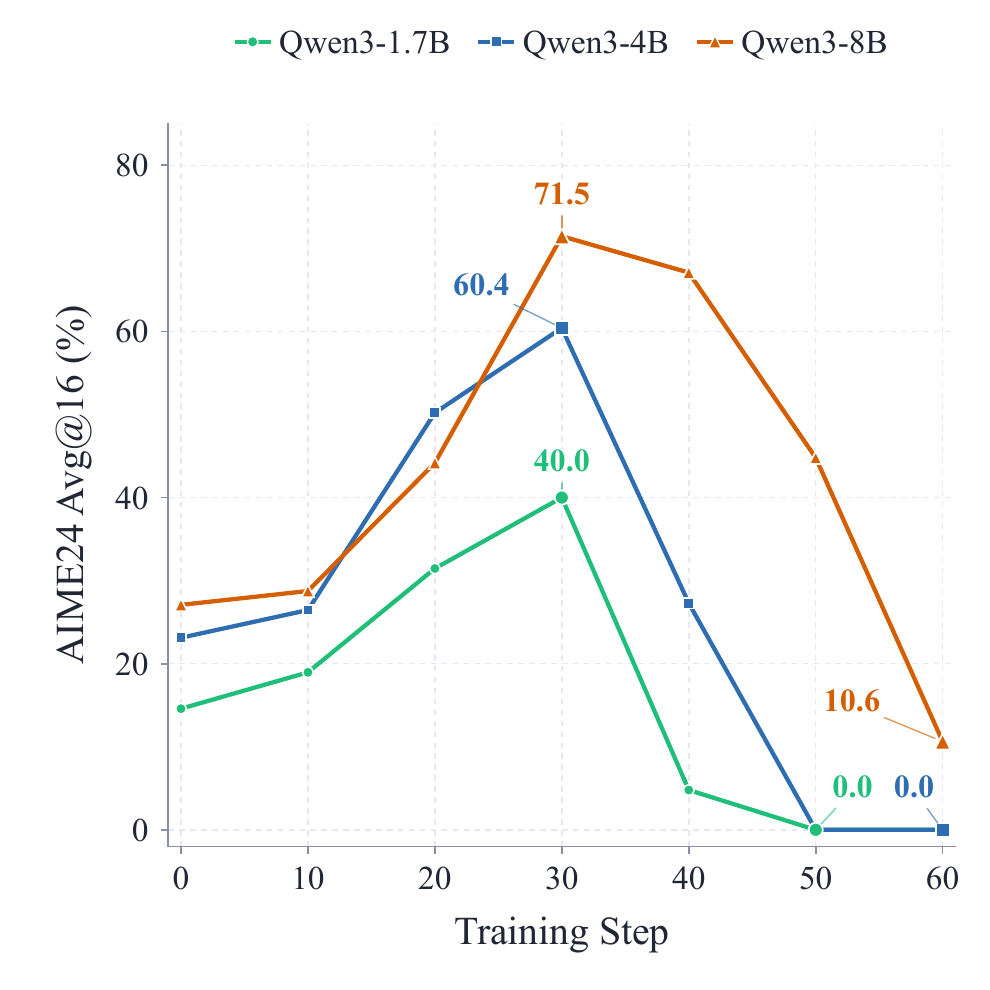}
        \caption{Avg@16.}
    \end{subfigure}\hfill
    \begin{subfigure}[t]{0.24\textwidth}
        \centering
        \includegraphics[width=\linewidth]{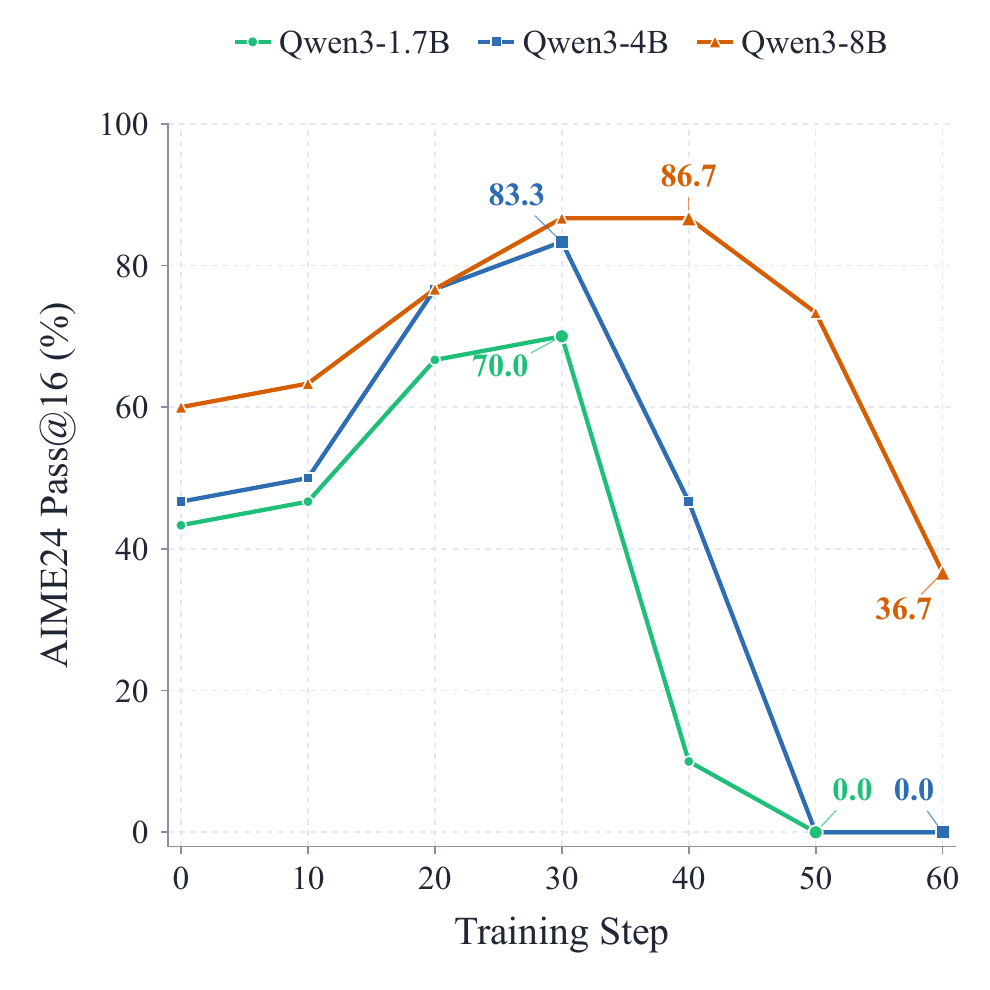}
        \caption{Pass@16.}
    \end{subfigure}
    \caption{TRA with irrelevant privileged solutions ($n=1$).}
    \label{fig:mismatch-solution}
\end{figure*}

\begin{figure}[t]
    \centering
    \includegraphics[width=\columnwidth]
        {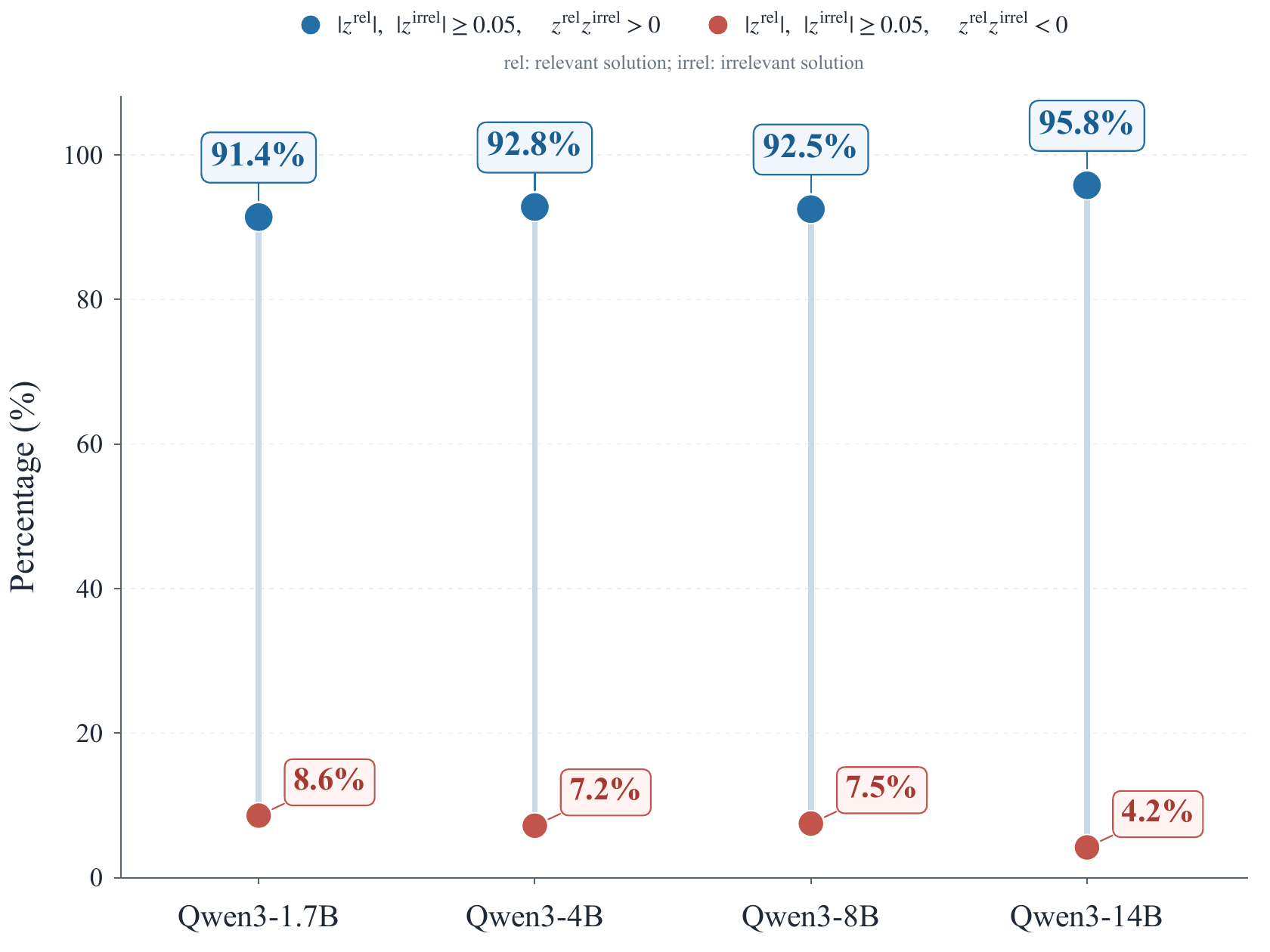}
    \caption{Direction agreement between relevant and irrelevant privileged solutions.}
    \label{fig:direction-agreement}
\end{figure}

\subsection{Beyond Correct and Relevant Solutions}

To examine whether privileged supervision requires relevant solutions, we construct an irrelevant-solution setting. For each training problem, we randomly sample a solution from another example in the training set and provide it as privileged information to the teacher, while the student receives only the original problem. This setting allows us to investigate the effects of irrelevant privileged solutions on reasoning behavior and performance.

As shown in Fig.~\ref{fig:mismatch-solution}, TRA with irrelevant privileged solutions exhibits similar behavior to that with relevant solutions. Across all three models, student--teacher KL divergence and reasoning length increase substantially, while Avg@16 reaches 40.0\%, 60.4\%, and 71.5\% after 30 steps, respectively, substantially outperforming OPSD baselines. Performance subsequently declines as responses approach the 32K generation limit and become truncated. Remarkably, even completely irrelevant solutions induce essentially the same reasoning expansion and performance gains as relevant solutions, demonstrating that the benefits of moving away from privileged teachers do not depend on semantic relevance.

To investigate how relevant and irrelevant privileged solutions affect token-level likelihoods, we adopt the likelihood-shift analysis protocol described in~\citep{he2026not,he2026calibrating} and use Qwen3-1.7B, Qwen3-4B, Qwen3-8B, and Qwen3-14B to compare their likelihood shifts over 4,000 rollouts. Let $x$ denote the input problem, $y=(y_1,\ldots,y_T)$ denote a rollout, and $y_{<i}$ its preceding tokens. We denote the relevant and irrelevant privileged solutions by $s^{\mathrm{rel}}$ and $s^{\mathrm{irrel}}$, respectively. For each token $y_i$, we define the corresponding likelihood shifts as
\begin{equation}
\begin{aligned}
z_i^{\mathrm{rel}}
&=
\log \pi_\theta(y_i \mid x,s^{\mathrm{rel}},y_{<i})
-
\log \pi_\theta(y_i \mid x,y_{<i}),\\
z_i^{\mathrm{irrel}}
&=
\log \pi_\theta(y_i \mid x,s^{\mathrm{irrel}},y_{<i})
-
\log \pi_\theta(y_i \mid x,y_{<i}).
\end{aligned}
\label{eq:privileged-likelihood-shift}
\end{equation}

\begin{figure}[t]
    \centering
    \includegraphics[width=\columnwidth]
        {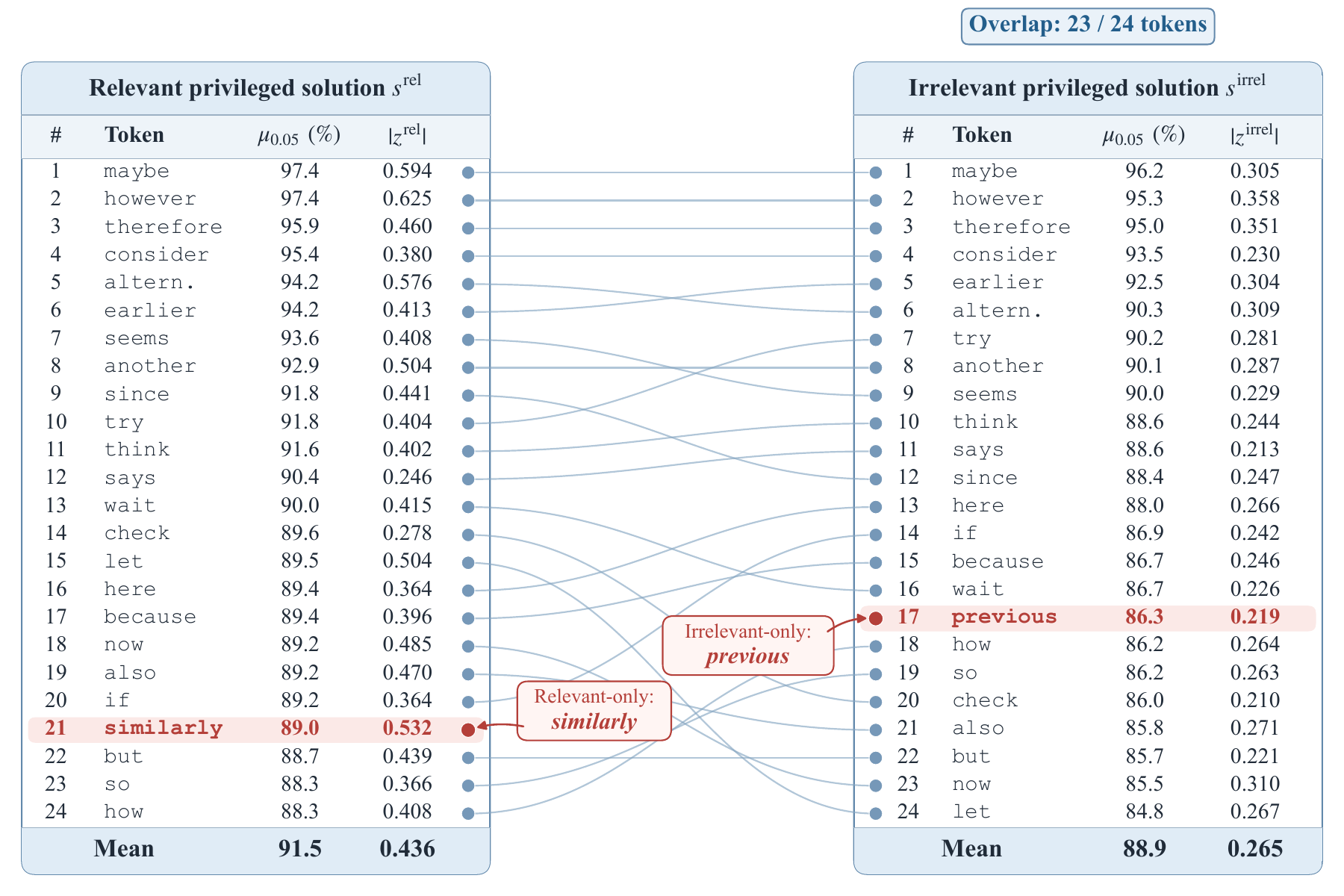}
    \caption{Token-level likelihood shifts under relevant and irrelevant privileged solutions.}
    \label{fig:token-privilege-comparison}
\end{figure}

\begin{figure*}[t]
    \centering
    \includegraphics[width=0.8\textwidth]
        {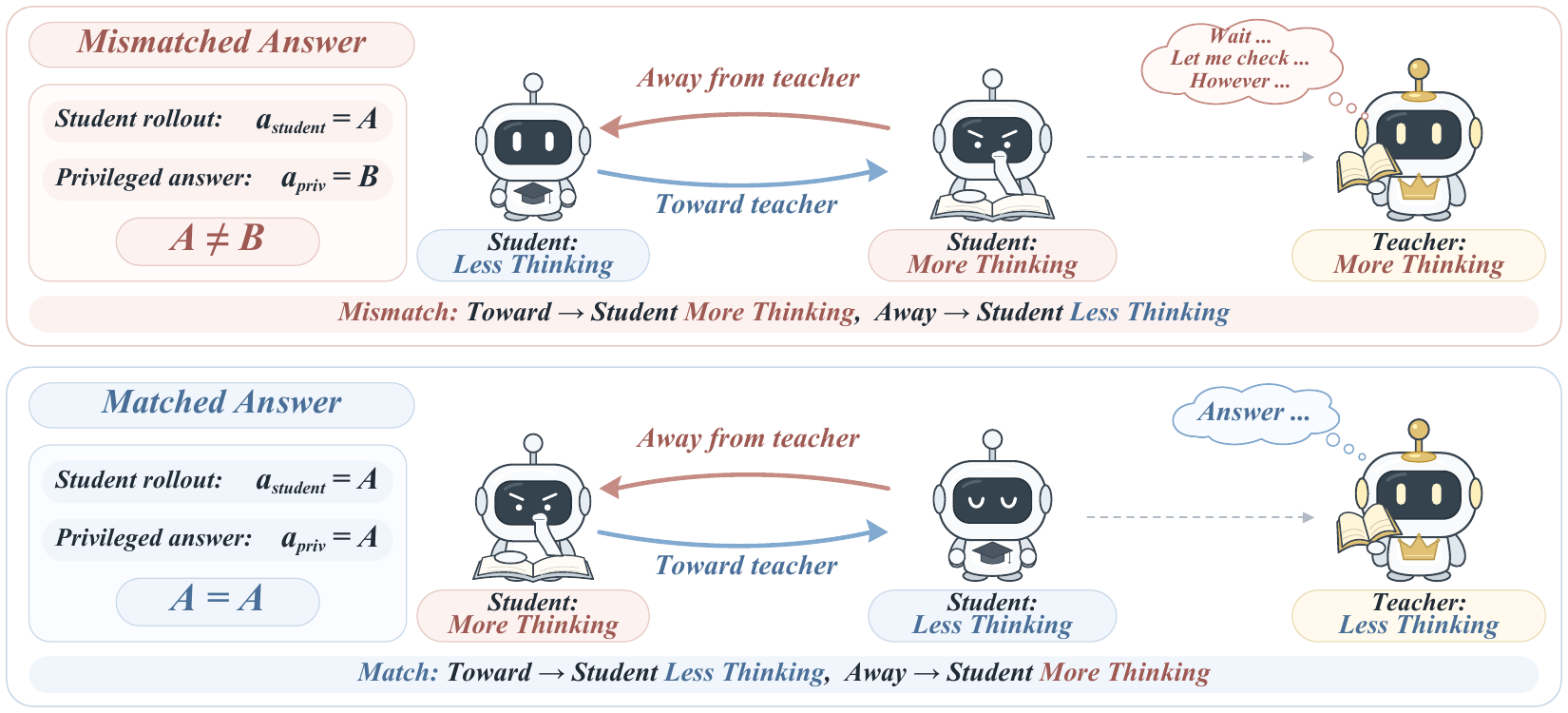}
    \caption{Thinking control under matched and mismatched privileged answers.}
    \label{fig:answer-consistency}
\end{figure*}

As shown in Fig.~\ref{fig:direction-agreement}, the likelihood shifts induced by relevant and irrelevant privileged solutions exhibit remarkably consistent directions across all four models. Among tokens for which both likelihood shifts exceed 0.05 in absolute magnitude, over 91\% exhibit the same direction, reaching 91.4\%, 92.8\%, 92.5\%, and 95.8\% for Qwen3-1.7B, 4B, 8B, and 14B, respectively. Opposite-direction shifts account for only a small minority. This strong directional agreement demonstrates that even semantically irrelevant privileged solutions induce nearly identical token-level distributional shifts to relevant solutions, further supporting our hypothesis that privileged teachers primarily regulate reasoning behavior rather than transfer solution-specific knowledge.

For each token category, we record its occurrence count, average likelihood shift, and the high-shift frequency $\mu_{0.05}$, defined as the proportion of occurrences with $|z_i|>0.05$:
\begin{equation}
    \mu_{0.05}^{(c)}
    =
    \frac{1}{|\mathcal I_c|}
    \sum_{i\in\mathcal I_c}
    \mathbb{I}(|z_i|>0.05),
\end{equation}
where $\mathcal I_c$ denotes the occurrences of token category $c$.

As shown in Fig.~\ref{fig:token-privilege-comparison}, tokens with substantial likelihood shifts are overwhelmingly dominated by reasoning-related expressions, including ``maybe'', ``however'', ``therefore'', ``consider'', ``wait'', and ``check''. These tokens exhibit remarkably high $\mu_{0.05}$ values, averaging 91.5\% and 88.9\% under relevant and irrelevant privileged solutions, respectively. More strikingly, 23 out of the top 24 tokens overlap between the two settings. Combined with the strong directional agreement in Fig.~\ref{fig:direction-agreement}, these results demonstrate that relevant and irrelevant privileged solutions induce highly similar token-level likelihood shifts. In both cases, privileged teachers predominantly suppress exploratory and reflective reasoning, explaining why moving away from either type of teacher can activate students' reasoning capabilities.

\begin{keynote}{Finding 5: Irrelevant Privilege}
\textbf{Irrelevant solutions are also thinking switches.}

They induce similar less-thinking biases, and moving away activates reasoning and improves performance.
\end{keynote}

\begin{figure*}[t]
    \centering
    \begin{subfigure}[t]{0.24\textwidth}
        \centering
        \includegraphics[width=\linewidth]{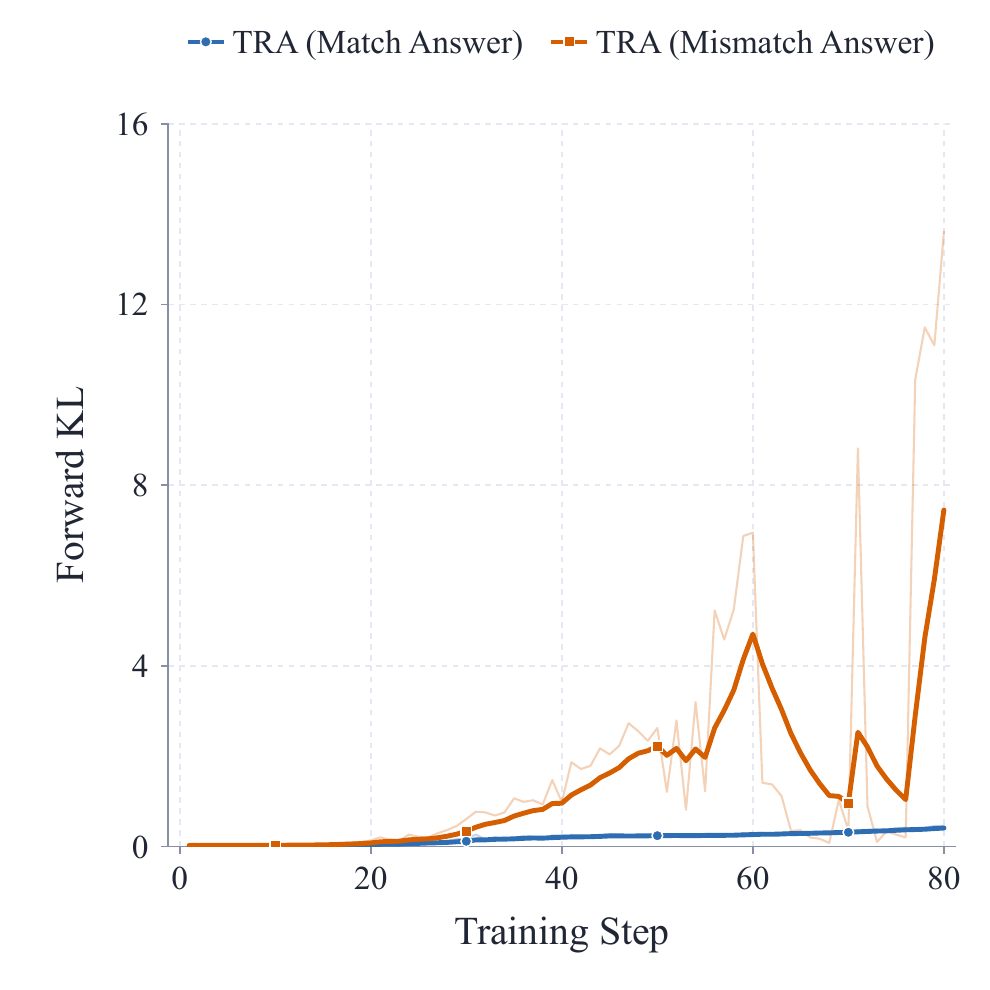}
        \caption{Student--teacher KL.}
    \end{subfigure}\hfill
    \begin{subfigure}[t]{0.24\textwidth}
        \centering
        \includegraphics[width=\linewidth]{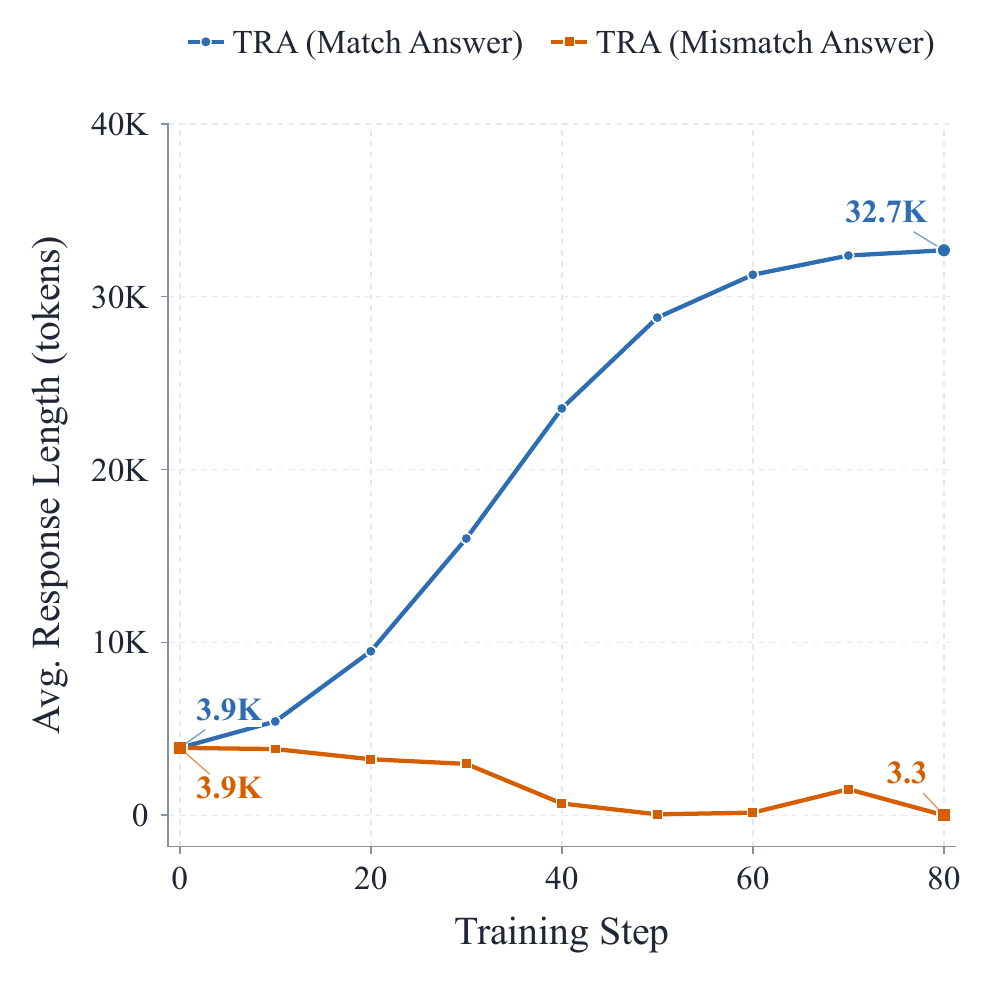}
        \caption{Reasoning length.}
    \end{subfigure}\hfill
    \begin{subfigure}[t]{0.24\textwidth}
        \centering
        \includegraphics[width=\linewidth]{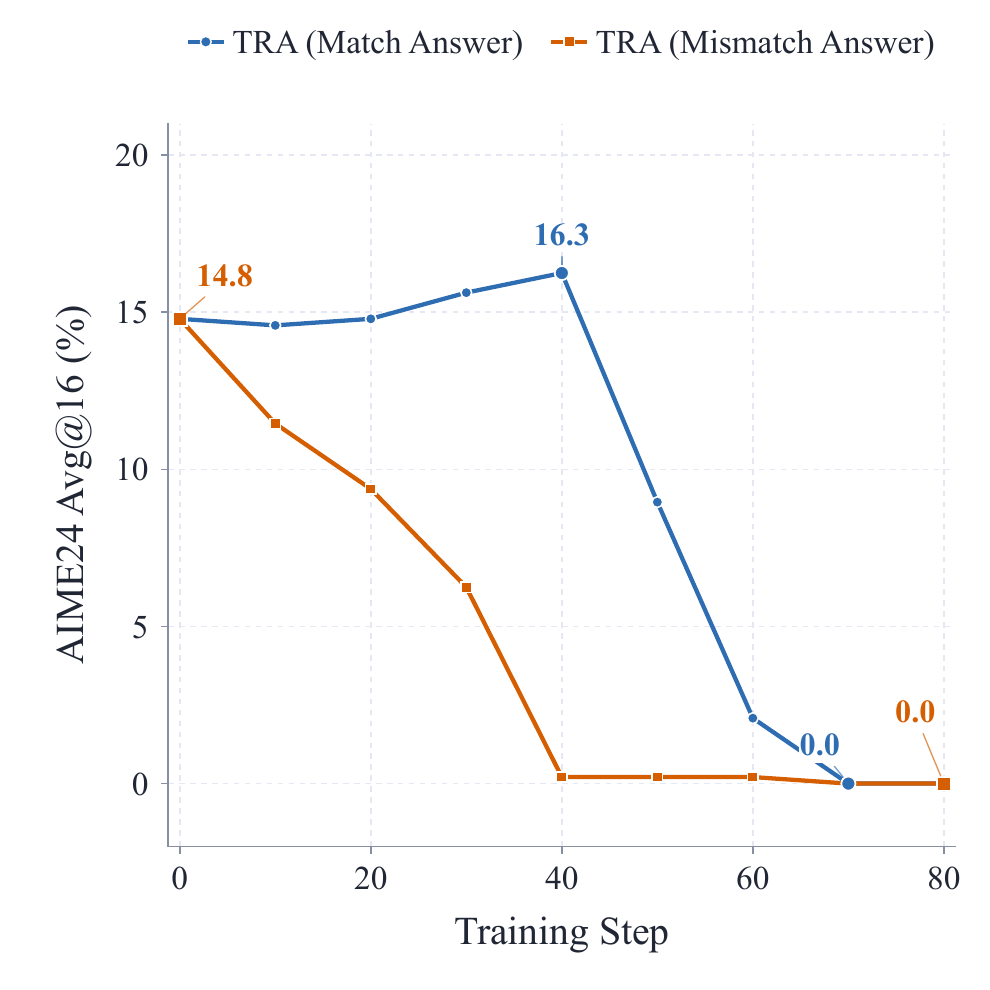}
        \caption{Avg@16.}
    \end{subfigure}\hfill
    \begin{subfigure}[t]{0.24\textwidth}
        \centering
        \includegraphics[width=\linewidth]{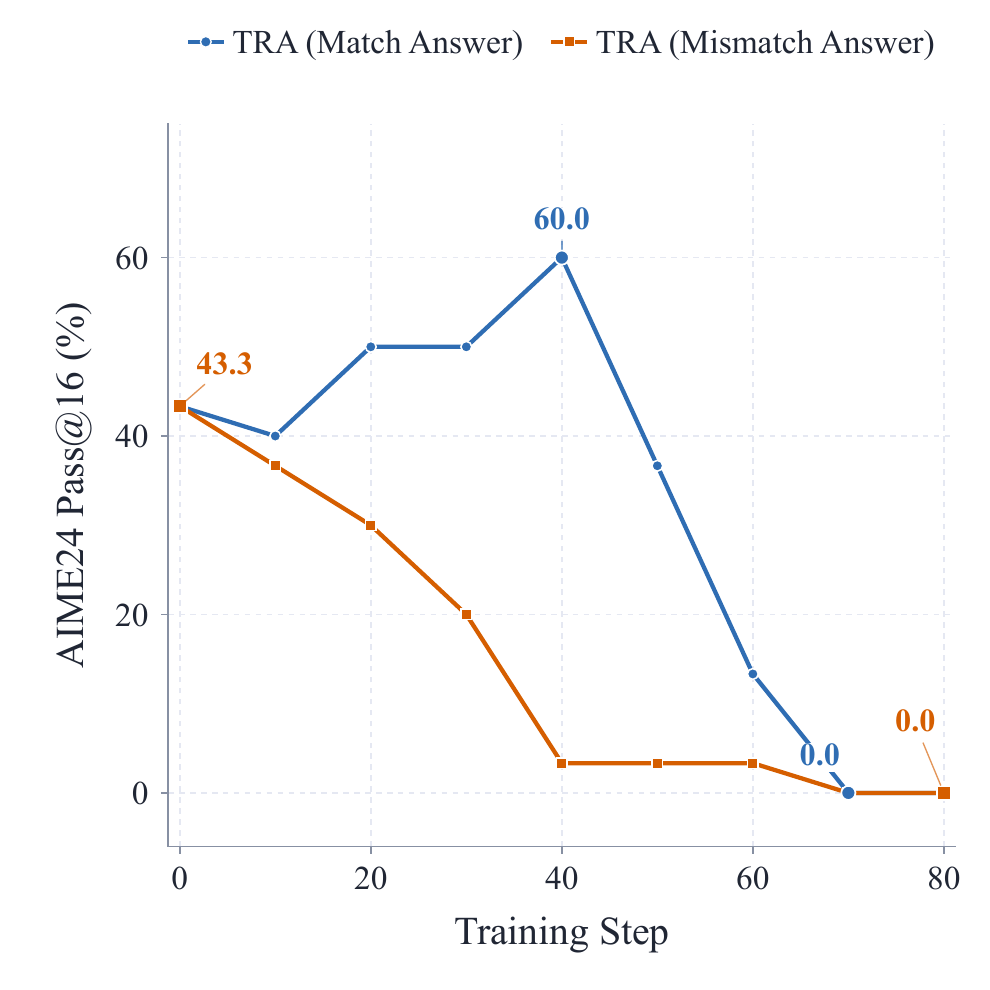}
        \caption{Pass@16.}
    \end{subfigure}
    \caption{TRA with matched and mismatched privileged answers. ($n=1$).}
    \label{fig:privi-answer}
\end{figure*}

\subsection{Answer-Only Privileged Supervision}

We investigate how answer-only privileged information influences thinking control. Unlike solution-based privilege, answer-only supervision does not exhibit a consistent tendency to increase or decrease reasoning length. \textbf{We hypothesize that its effect depends on the consistency between the privileged answer and the answer generated by the student's trajectory.} As illustrated in Fig.~\ref{fig:answer-consistency}, when the two answers mismatch, the privileged teacher tends to exhibit more extensive reasoning, such that moving toward the teacher induces more thinking, whereas moving away induces less. Conversely, when they match, the teacher already has access to the answer reached by the student and tends to exhibit less thinking. Consequently, moving toward the teacher suppresses reasoning, whereas moving away encourages more extensive thinking.

As shown in Fig.~\ref{fig:privi-answer}, TRA induces strikingly opposite reasoning behaviors under matched and mismatched answer supervision, consistent with our hypothesis. With Match Answer, reasoning length continuously expands from 3.9K to 32.7K tokens over 80 optimization steps. In stark contrast, with Mismatch Answer, \textbf{the average response length collapses from 3.9K to merely 3.3 tokens by step 80, effectively eliminating reasoning altogether.} These results demonstrate that reversing the teacher alignment direction produces opposite effects depending on answer--trajectory consistency. Notably, for Qwen3-1.7B, TRA with Match Answer also improves both Avg@16 and Pass@16, from 14.8\% to 16.3\% and from 43.3\% to 60.0\%, respectively. Overall, these contrasting behaviors strongly support our hypothesis that answer--trajectory consistency determines the direction of thinking control.

\begin{keynote}{Finding 6: Answer--Trajectory Consistency}
\textbf{Consistency matters more than correctness.}

Answer--trajectory consistency determines whether privileged alignment promotes or suppresses student reasoning.
\end{keynote}

\subsection{Distinct Mechanisms of Thinking Control}

Despite using the same TRA objective, solution-based and answer-only privileged supervision exhibit substantially different effects on student reasoning. While solution-based privilege consistently induces pronounced reasoning expansion and significant performance gains, answer-only privilege produces markedly different, answer-consistency-dependent reasoning behaviors, with little corresponding performance improvement. This striking discrepancy motivates a closer examination of the token-level effects induced by these two types of privileged information.

As shown in Table~\ref{tab:distinct-thinking-control}, privileged solutions substantially reduce the likelihood of reasoning-control tokens such as ``wait'' and ``check'', suggesting that teachers tend to take reasoning shortcuts. TRA encourages students to avoid these shortcuts and reason more extensively. In contrast, privileged answers exhibit a distinctive effect on reasoning termination: matched answers increase the EOS likelihood, whereas mismatched answers decrease it. Thus, TRA with matched answers encourages students to continue reasoning while exerting a much weaker effect on reasoning-control tokens. As shown in Table~\ref{tab:solution-answer-comparison}, although both settings induce more thinking, solution-based TRA achieves 39.6\% Avg@16, far exceeding the 16.3\% achieved with matched answers. \textbf{Different privileged information induces more thinking through distinct mechanisms.}

\begin{table}[t]
    \centering
    \caption{Token-level likelihood shifts under different privileges.}
    \label{tab:distinct-thinking-control}
    \small
    \setlength{\tabcolsep}{6pt}
    \renewcommand{\arraystretch}{1.25}
    \begin{tabular}{lrrr}
        \toprule
        & \textbf{Solution}
        & \multicolumn{2}{c}{\textbf{Answer Privilege}} \\
        \cmidrule(lr){2-2}
        \cmidrule(lr){3-4}
        \textbf{Token}
        & \textbf{Privilege}
        & \textbf{Match}
        & \textbf{Mismatch} \\
        \midrule

        \multicolumn{4}{l}{
            \textit{\footnotesize Reasoning-control tokens}} \\

        \texttt{wait}
        & \cellcolor{reasonblue}$-1.039$
        & $-0.338$ & $-0.257$ \\

        \texttt{but}
        & \cellcolor{reasonblue}$-0.347$
        & $-0.196$ & $-0.065$ \\

        \texttt{may}
        & \cellcolor{reasonblue}$-0.206$
        & $-0.020$ & $-0.022$ \\

        \texttt{so}
        & \cellcolor{reasonblue}$-0.431$
        & $-0.088$ & $-0.069$ \\

        \texttt{check}
        & \cellcolor{reasonblue}$-0.250$
        & $-0.087$ & $-0.073$ \\

        \midrule

        \multicolumn{4}{l}{
            \textit{\footnotesize Termination-control token}} \\

        \textbf{EOS}
        & $-0.110$
        & \cellcolor{stoporange}$\mathbf{+0.026}$
        & \cellcolor{stoporange}$\mathbf{-0.230}$ \\

        \bottomrule
    \end{tabular}
\end{table}

\begin{table*}[!t]
    \centering
    \caption{
        Avg@16 (\%) across seven reasoning benchmarks
        in non-thinking mode.
        The student and teacher are Qwen3-1.7B and
        Qwen3-4B-Instruct-2507, respectively.
        PI denotes a privileged solution.
        $\downarrow$ indicates decreases relative to OPD
        without PI, in percentage points.
        Scores and decreases are rounded independently.
    }
    \label{tab:external_teacher_results}
    \small
    \setlength{\tabcolsep}{4pt}
    \renewcommand{\arraystretch}{1.20}
    \begin{tabular*}{\textwidth}{
        @{\extracolsep{\fill}}lrrrrrrrr@{}
    }
        \toprule
        \textbf{Setting}
        & \textbf{AMC23}
        & \textbf{AIME24}
        & \textbf{AIME25}
        & \textbf{AIME26}
        & \textbf{MATH-500}
        & \textbf{SciBench}
        & \textbf{LogiQA2.0}
        & \textbf{Mean} \\
        \midrule

        Student
        & 46.7 & 14.4 & 6.9 & 8.8
        & 76.6 & 37.5 & 58.5 & 35.6 \\
        Teacher
        & 93.3 & 62.7 & 44.2 & 52.1
        & 96.0 & 63.7 & 80.5 & 70.4 \\

        \midrule

        OPD w/o PI (FKL)
        & 76.9 & 40.6 & 28.5 & 27.3
        & 89.7 & 47.9 & 64.1 & \textbf{53.6} \\
        \textbf{OPD w/ PI}
        & 73.6\textsubscript{$\downarrow 3.3$}
        & 36.7\textsubscript{$\downarrow 4.0$}
        & 29.4
        & 21.3\textsubscript{$\downarrow 6.0$}
        & 88.0\textsubscript{$\downarrow 1.6$}
        & 45.5\textsubscript{$\downarrow 2.4$}
        & 64.0\textsubscript{$\downarrow 0.1$}
        & \textbf{51.2}\textsubscript{$\downarrow 2.4$} \\

        \bottomrule
    \end{tabular*}
\end{table*}

\begin{table}[t]
    \centering
    \caption{Comparison of solution-based and answer-only TRA on AIME24 (Qwen3-1.7B).}
    \label{tab:solution-answer-comparison}
    \small
    \setlength{\tabcolsep}{3pt}
    \begin{tabular}{lcccc}
        \toprule
        Method & Initial & Best & $\Delta$ & Step \\
        \midrule
        TRA (Solution) & 14.4 & \textbf{39.6} & \textbf{+25.2} & 40 \\
        TRA (Match Answer) & 14.4 & 16.3 & +1.9 & 40 \\
        \bottomrule
    \end{tabular}
\end{table}

\begin{keynote}{Finding 7: Distinct Thinking Mechanisms}
\textbf{More thinking does not guarantee better performance.}

Different types of privileged information control thinking through distinct mechanisms, with varying effects on reasoning performance.
\end{keynote}

\begin{figure}[t]
    \centering
    \begin{subfigure}{0.49\columnwidth}
        \centering
        \includegraphics[width=\linewidth]
            {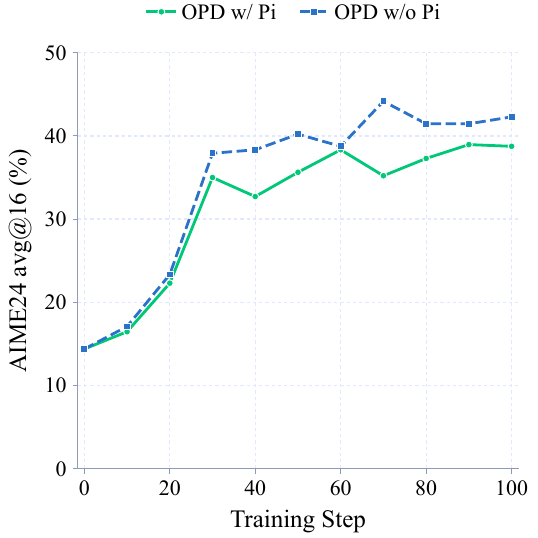}
        \caption{Avg@16}
    \end{subfigure}
    \hfill
    \begin{subfigure}{0.49\columnwidth}
        \centering
        \includegraphics[width=\linewidth]
            {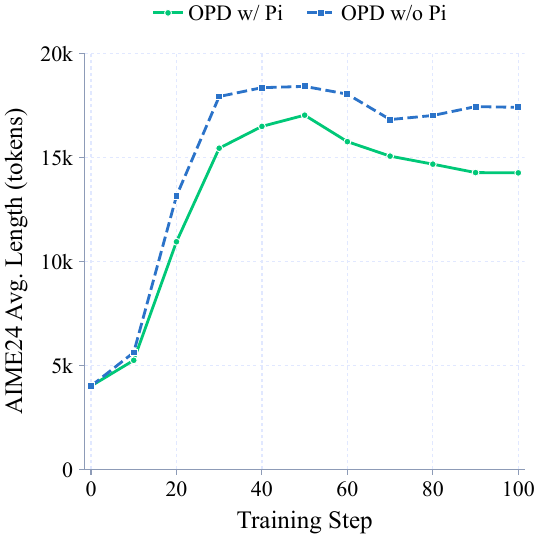}
        \caption{Response Length}
    \end{subfigure}
    \caption{
        Effects of privileged information (PI) on OPD.
        PI denotes a verified solution.
        The student and teacher are Qwen3-1.7B and
        Qwen3-4B-Instruct-2507, respectively.
        Both operate in non-thinking mode.
    }
    \label{fig:external-teacher-pi}
\end{figure}

\subsection{Privileged Information Can Weaken Strong Teachers}

To investigate whether privileged information benefits distillation from stronger teachers, we employ Qwen3-4B-Instruct-2507 as the teacher and Qwen3-1.7B~\citep{yang2025qwen3} as the student, both operating in non-thinking mode. We adopt the forward-KL objective and compare two settings: standard OPD, where the teacher receives only the problem, and privileged OPD, where the teacher additionally receives the reference solution. All other training configurations remain identical.

As shown in Fig.~\ref{fig:external-teacher-pi}, although both OPD settings substantially improve student performance, introducing privileged solutions consistently leads to lower Avg@16 and shorter reasoning lengths throughout training compared with standard OPD. This degradation further extends across diverse reasoning benchmarks. As shown in Table~\ref{tab:external_teacher_results}, introducing privileged solutions reduces mean Avg@16 from 53.6\% to 51.2\%, with performance degradation on six of seven benchmarks. Notably, performance on AIME24 and AIME26 decreases by 4.0 and 6.0 percentage points, respectively. These observations are consistent with our findings in self-distillation: privileged solutions suppress student thinking through distributional alignment, thereby weakening the benefits of distillation. \textbf{Even when the teacher is substantially stronger than the student, providing privileged solutions suppresses reasoning and compromises distillation effectiveness.} These findings further support our interpretation of privileged information as a thinking-control signal rather than a source of knowledge to be distilled.

\begin{keynote}{Finding 8: Strong Teachers}
\textbf{Privileged information weakens strong teachers.}

Privileged information suppresses the reasoning of strong teachers, thereby reducing their effectiveness in enhancing student reasoning.
\end{keynote}

\begin{table*}[!t]
    \centering
    \caption{
        Avg@16 (\%) across seven reasoning benchmarks
        in non-thinking mode.
        The best result for each model size is highlighted.
        Subscripts indicate gains over Base in percentage points.
        Scores and gains are rounded independently.
    }
    \label{tab:main_avg16}
    \small
    \setlength{\tabcolsep}{4pt}
    \renewcommand{\arraystretch}{1.20}
    \begin{tabular*}{\textwidth}{
        @{\extracolsep{\fill}}llrrrrrrrr@{}
    }
        \toprule
        \textbf{Model} & \textbf{Method}
        & \textbf{AMC23}
        & \textbf{AIME24}
        & \textbf{AIME25}
        & \textbf{AIME26}
        & \textbf{MATH-500}
        & \textbf{SciBench}
        & \textbf{LogiQA2.0}
        & \textbf{Mean} \\
        \midrule

        \multirow{3}{*}{\textbf{Qwen3-1.7B}}
        & Base
        & 46.7 & 14.4 & 6.9 & 8.8
        & 76.6 & 37.5 & 58.5 & 35.6 \\
        & OPSD
        & 49.8 & 18.3 & 15.6 & 16.9
        & 78.6 & 20.6 & 57.2 & 36.7 \\
        & TRA
        & \bestscore{70.6}\textsubscript{+23.9}
        & \bestscore{41.0}\textsubscript{+26.7}
        & \bestscore{31.3}\textsubscript{+24.4}
        & \bestscore{31.3}\textsubscript{+22.5}
        & \bestscore{90.1}\textsubscript{+13.5}
        & \bestscore{48.6}\textsubscript{+11.0}
        & \bestscore{60.4}\textsubscript{+1.8}
        & \bestscore{53.3}\textsubscript{+17.7} \\

        \midrule

        \multirow{3}{*}{\textbf{Qwen3-4B}}
        & Base
        & 66.6 & 23.1 & 19.2 & 17.3
        & 86.9 & 51.6 & 68.6 & 47.6 \\
        & OPSD
        & 72.7 & 27.5 & 25.8 & 24.0
        & 88.8 & 52.4 & 70.5 & 51.7 \\
        & TRA
        & \bestscore{95.0}\textsubscript{+28.4}
        & \bestscore{67.1}\textsubscript{+44.0}
        & \bestscore{61.0}\textsubscript{+41.9}
        & \bestscore{61.7}\textsubscript{+44.4}
        & \bestscore{95.8}\textsubscript{+8.9}
        & \bestscore{67.0}\textsubscript{+15.4}
        & \bestscore{75.0}\textsubscript{+6.4}
        & \bestscore{74.7}\textsubscript{+27.0} \\

        \midrule

        \multirow{3}{*}{\textbf{Qwen3-8B}}
        & Base
        & 69.7 & 27.1 & 19.4 & 15.0
        & 87.3 & 55.8 & 73.9 & 49.7 \\
        & OPSD
        & 82.5 & 48.5 & 37.9 & 37.9
        & 92.6 & 59.2 & 76.5 & 62.2 \\
        & TRA
        & \bestscore{95.3}\textsubscript{+25.6}
        & \bestscore{73.8}\textsubscript{+46.7}
        & \bestscore{60.4}\textsubscript{+41.0}
        & \bestscore{63.1}\textsubscript{+48.1}
        & \bestscore{96.1}\textsubscript{+8.7}
        & \bestscore{66.6}\textsubscript{+10.9}
        & \bestscore{79.4}\textsubscript{+5.5}
        & \bestscore{76.4}\textsubscript{+26.7} \\

        \bottomrule
    \end{tabular*}
\end{table*}

\begin{table*}[!t]
    \centering
    \caption{
        Pass@16 (\%) across seven reasoning benchmarks
        in non-thinking mode.
        The best result for each model size is highlighted.
        Subscripts indicate differences from Base in percentage points.
        Scores and differences are rounded independently.
    }
    \label{tab:main_pass16}
    \small
    \setlength{\tabcolsep}{4pt}
    \renewcommand{\arraystretch}{1.20}
    \begin{tabular*}{\textwidth}{
        @{\extracolsep{\fill}}llrrrrrrrr@{}
    }
        \toprule
        \textbf{Model} & \textbf{Method}
        & \textbf{AMC23}
        & \textbf{AIME24}
        & \textbf{AIME25}
        & \textbf{AIME26}
        & \textbf{MATH-500}
        & \textbf{SciBench}
        & \textbf{LogiQA2.0}
        & \textbf{Mean} \\
        \midrule

        \multirow{3}{*}{\textbf{Qwen3-1.7B}}
        & Base
        & 82.5 & 40.0 & 16.7 & 26.7
        & 94.0 & 66.0 & 84.5 & 58.6 \\
        & OPSD
        & 87.5 & 53.3 & 43.3 & 43.3
        & 94.5 & 49.5 & 82.5 & 64.9 \\
        & TRA
        & \bestscore{92.5}\textsubscript{+10.0}
        & \bestscore{73.3}\textsubscript{+33.3}
        & \bestscore{63.3}\textsubscript{+46.7}
        & \bestscore{53.3}\textsubscript{+26.7}
        & \bestscore{98.5}\textsubscript{+4.5}
        & \bestscore{76.0}\textsubscript{+10.0}
        & \bestscore{88.5}\textsubscript{+4.0}
        & \bestscore{77.9}\textsubscript{+19.3} \\

        \midrule

        \multirow{3}{*}{\textbf{Qwen3-4B}}
        & Base
        & 95.0 & 46.7 & 43.3 & 40.0
        & 98.0 & 72.5 & \bestscore{87.5} & 69.0 \\
        & OPSD
        & 95.0 & 66.7 & 60.0 & 43.3
        & 96.0 & 70.5 & \bestscore{87.5} & 74.1 \\
        & TRA
        & \bestscore{100.0}\textsubscript{+5.0}
        & \bestscore{90.0}\textsubscript{+43.3}
        & \bestscore{80.0}\textsubscript{+36.7}
        & \bestscore{86.7}\textsubscript{+46.7}
        & \bestscore{100.0}\textsubscript{+2.0}
        & \bestscore{81.5}\textsubscript{+9.0}
        & 86.5\textsubscript{-1.0}
        & \bestscore{89.2}\textsubscript{+20.2} \\

        \midrule

        \multirow{3}{*}{\textbf{Qwen3-8B}}
        & Base
        & 95.0 & 60.0 & 43.3 & 43.3
        & 98.5 & 77.0 & \bestscore{93.0} & 72.9 \\
        & OPSD
        & \bestscore{100.0} & 80.0 & 66.7 & 76.7
        & \bestscore{99.5} & 76.0 & 92.0 & 84.4 \\
        & TRA
        & \bestscore{100.0}\textsubscript{+5.0}
        & \bestscore{93.3}\textsubscript{+33.3}
        & \bestscore{80.0}\textsubscript{+36.7}
        & \bestscore{86.7}\textsubscript{+43.3}
        & \bestscore{99.5}\textsubscript{+1.0}
        & \bestscore{81.5}\textsubscript{+4.5}
        & 91.5\textsubscript{-1.5}
        & \bestscore{90.4}\textsubscript{+17.5} \\

        \bottomrule
    \end{tabular*}
\end{table*}

\section{Reasoning Performance and Cross-Domain Generalization}

To investigate whether TRA activates general reasoning capabilities beyond the training domain, we select the best-performing checkpoints of OPSD and TRA for each model and evaluate them across seven benchmarks spanning three reasoning domains: mathematical reasoning on AMC23, AIME24, AIME25, AIME26, and MATH-500~\citep{lightman2024let}; scientific reasoning on SciBench~\citep{wang2023scibench}; and logical reasoning on LogiQA2.0~\citep{liu2023logiqa}. For benchmarks containing more than 200 examples, we uniformly sample 200 examples at random to accelerate evaluation. Notably, all models are trained exclusively on mathematical problems, allowing us to examine whether the reasoning gains generalize beyond mathematics.

As shown in Tables~\ref{tab:main_avg16} and~\ref{tab:main_pass16}, TRA demonstrates remarkable effectiveness across all three model scales, outperforming both Base and OPSD in Avg@16 across all 21 model--benchmark comparisons and achieving the best or tied-best Pass@16 in 19 out of 21 comparisons. Compared with OPSD, TRA improves mean Avg@16 by 16.6, 23.0, and 14.2 percentage points and mean Pass@16 by 13.0, 15.1, and 6.0 percentage points for Qwen3-1.7B, 4B, and 8B, respectively.

Remarkably, despite being trained exclusively on mathematical problems, TRA improves Avg@16 over the base models on SciBench by 11.0, 15.4, and 10.9 percentage points, and on LogiQA2.0 by 1.8, 6.4, and 5.5 percentage points, respectively. These consistent improvements across scientific and logical reasoning demonstrate that the reasoning capabilities activated by TRA generalize beyond the training domain. Nevertheless, the improvements in non-mathematical reasoning are generally smaller than those on competition-level mathematical benchmarks, indicating that the benefits of enhanced thinking remain partially domain-dependent.

\section{Discussion}

Our findings fundamentally challenge the knowledge-distillation interpretation of OPSD. Direct alignment with solution-privileged teachers suppresses reasoning and degrades performance, whereas moving away from them substantially improves reasoning capabilities, even when the privileged solutions are completely irrelevant. These observations support a unified interpretation: \textbf{privileged information does not serve as knowledge to be distilled, but as a signal that controls thinking.}

Moreover, different types of privileged information induce distinct changes in the teacher's reasoning behavior. Privileged solutions primarily influence reasoning patterns, whereas privileged answers primarily regulate reasoning termination. Consequently, different forms of thinking control can lead to substantially different performance outcomes, even when they all encourage more thinking. We therefore argue that future research on privileged supervision should shift its focus from how to better distill privileged teachers to how different privileged contexts shape teacher reasoning and how the resulting distributional shifts can be exploited to control student thinking.

\section{Conclusion}

In this work, we revisit the fundamental role of privileged teachers in OPSD and challenge its conventional interpretation as knowledge distillation. Through theoretical analysis and extensive experiments, we demonstrate that aligning with privileged teachers can suppress reasoning, whereas reversing the alignment can activate substantial reasoning capabilities, even when the privileged solutions are irrelevant. We further reveal that different types of privileged information control thinking through distinct mechanisms, and that more thinking does not necessarily translate into better performance. The effectiveness and cross-domain generalization of TRA further support our interpretation of privileged supervision as thinking control rather than knowledge transfer. Together, these findings lead to our central conclusion: \textbf{privileged teachers were never about teaching knowledge; they were thinking switches all along.}

% Research sections below were empty in the supplied manuscript.
% Retained as source scaffolding; uncomment each heading when its text is ready.
% \subsection{Teacher Reversal Alignment}
% \subsection{Thinking Control Beyond Knowledge Transfer}
% \section{Thinking-Controlled Policy Optimization}
% \section{Discussion}
% \section{Conclusion}

\bibliographystyle{icml2027}
\bibliography{reference}

% Add an appendix here when substantive appendix material is available.
% \appendix
% \section{Additional Details}
\end{document}